%% file: iclr2026_conference.tex
\documentclass{article} 
\usepackage{iclr2026_conference,times}

\input{math_commands.tex}

\usepackage{hyperref}
\usepackage{url}

\usepackage[utf8]{inputenc} 
\usepackage[T1]{fontenc}    
\usepackage{hyperref}       
\usepackage{url}            
\usepackage{booktabs}       
\usepackage{amsmath,amsfonts}
\usepackage{nicefrac}
\usepackage{microtype}
\usepackage{xcolor}
\usepackage{soul}
\usepackage{graphicx}
\usepackage{subcaption}
\usepackage{algorithm}
\usepackage[noend]{algpseudocode}
\usepackage{listings}

\title{Enter the Void: Exploring with High Entropy Plans}

\author{Ashish Sundar \\
Department of Computer Science \\
University of Exeter \\
\texttt{as1748@exeter.ac.uk}
\And
Chunbo Luo \\
Department of Computer Science \\
University of Exeter \\
\texttt{c.luo@exeter.ac.uk}
\And
Xiaoyang Wang \\
Department of Computer Science \\
University of Exeter \\
\texttt{x.wang7@exeter.ac.uk}
}

\newcommand{\rev}[1]{\textcolor{blue}{#1}}
\renewcommand{\rev}[1]{#1}
\iclrfinalcopy 
\begin{document}

\maketitle

\begin{abstract}
  Model-based reinforcement learning (MBRL) offers an intuitive way to increase the sample efficiency of model-free RL methods by simultaneously training a world model that learns to predict the future. These models constitute the large majority of training compute and time and they are subsequently used to train actors entirely in simulation, but once this is done they are quickly discarded. \rev{We show in this work that utilising these models at inference time can significantly boost sample efficiency}. We propose a novel approach that anticipates and actively seeks out informative states using the world model’s short-horizon latent predictions, offering a principled alternative to traditional curiosity-driven methods that chase \rev{outdated estimates of high uncertainty states}. While many model predictive control (MPC) based methods offer similar alternatives, they typically lack commitment, synthesising multiple multi-step plans at every step. To mitigate this, we present a hierarchical planner that dynamically decides when to replan, planning horizon length, and the commitment to searching entropy. While our method can theoretically be applied to any model that trains its own actors with solely model generated data, we have applied it to Dreamer to illustrate the concept. Our method finishes MiniWorld's procedurally generated mazes 50\% faster than base Dreamer at convergence and in only 60\% of the environment steps that base Dreamer's policy needs; it displays reasoned exploratory behaviour in Crafter, achieves the same reward as base Dreamer in a third of the steps; \rev{planning tends to improve sample efficieny on DeepMind Control tasks.}
\end{abstract}

\section{Introduction}

In recent years, reinforcement learning (RL) has achieved remarkable success across a variety of domains, from mastering Go \citep{mgo} to racing drones at high speed \citep{drone}. However, these successes often rely on dense reward signals and highly structured environments. In real-world applications such as autonomous navigation, exploration, and disaster response, rewards are sparse and environments are stochastic and partially observable. In these conditions, achieving efficient exploration and good sample efficiency remain an active research problem. 

\rev{
Curiosity-based methods typically treat novelty as an intrinsic reward bonus that is added to the environment reward and propagated through the same value- and policy-learning mechanisms as extrinsic reward~\citep{Pathak2017Curiosity,burda2018exploration,Bellemare2016Unifying,ostrovski2017count,badia2020never,Raileanu2020RIDE}. These mechanisms are derived under a stationary MDP assumption in which the reward function $r(s,a)$ is fixed over time~\citep{sutton2018reinforcement}, so once a bonus has been associated with a state(-action) distribution it is effectively treated as part of a persistent ``true'' value. In practice, however, novelty-based bonuses are deliberately non-stationary: prediction errors, pseudo-counts, and related signals decay as states are revisited~\citep{ostrovski2017count,Bellemare2016Unifying,mahankali2022novelty}, and it is common to normalise intrinsic rewards online to stabilise learning under this drift~\citep{Burda2019LargeScale,burda2018exploration}. This yields a robust low-frequency signal for long-term novelty seeking, but it remains retrospective: the bonus only reflects novelty after it has propagated through the value function. By contrast, anticipatory exploration methods based on short-horizon predictions of epistemic uncertainty explicitly plan toward states that are \emph{predicted} to be novel before they are experienced~\citep{shyam2019model,sekar2020planning,jarrett2023hindsight}; our approach falls into this anticipatory category.}

\rev{Instead of having to learn novelty, we instead propose to use model uncertainty. To do this we use model-based reinforcement learning (MBRL),} where models of the environment (world models \citep{world_models}) are concurrently trained alongside the actor to predict environment transitions. \rev{Usually, the model is used to train the actor and is not used at inference time. We use the model as a planner at inference time to quantify the distribution entropy of the model's predicted next state.}

\rev{Seeking the next most uncertain state can lead to greedily optimising for noise, while falling into local aleatoric optima, so we use a planner to choose from a series of reward motivated trajectories holistically. To do this, we leverage the world model by combining it with the greedy actor to generate a selection of high scoring rollouts, from which we choose the rollout whose predicted states have the highest entropy and predicted reward. To keep the method reactive, we introduce a meta planner that can terminate planned trajectories when plans become stale and generate a new plan.}

\rev{Thus,} we propose to augment Dreamer \citep{hafner2020mastering}, a prominent and efficient world model, with a planning mechanism to anticipate and seek informative states as they are about to occur to drive reasoned exploration. \rev{This can be applied to any model based reinforcement learning method that trains the actor exclusively with the model (instead of the environment). This method can also be applied in conjunction with curiosity based methods to obtain both long and short term novelty, though we do not focus on that in this work. In this work,} we also introduce a lightweight Proximal Policy Optimisation (PPO) \cite{schulman2017ppo} based hierarchical planner that dynamically decides when to commit to a pre-selected rollout and when to discard it to replan. \rev{Importantly, we do not modify Dreamer’s world-model or actor training objectives; all losses and KL coefficients remain identical to DreamerV3. Our contribution is an inference-time entropy-seeking planner that changes action selection and thus the replay distribution.}

Our contributions are:
\begin{itemize}
    \item \rev{Use an MBRL model as a planner at inference time to proactively target} high-entropy states \rev{multiple steps (as far as the maximum world model horizon - up to 15 in this work) in the future, supporting seeking delayed gratification via reasoned anticipatory exploration.}
    \item \rev{Reinterpret the KL objective in} world-model training as a min–max interaction that couples model learning with entropy-seeking exploration, improving information gain and sample efficiency.
    \item Introducing a reactive hierarchical planner that dynamically selects between committing to a plan and replanning based on new information received, {reducing dithering and improving efficiency through learned plan commitment.}
\end{itemize}

The rest of the paper is structured as follows. In Section~\ref{sec:related}, we review related work in intrinsic motivation, planning, and hierarchical RL. In Section~\ref{sec:preliminaries}, we provide background on Dreamer and its training formulation. Section~\ref{sec:method} introduces our entropy-seeking planner and reactive hierarchical policy. Section~\ref{sec:experiments} details our experimental setup, evaluates performance on MiniWorld's procedurally generated 3D maze environment, Crafter, and DMC's vision based control environment. Section~\ref{sec:Conclusion} concludes this work and outlines limitations.

\section{Related Work}
\label{sec:related}

\subsection{Intrinsic Reward}
Intrinsic motivation methods can be grouped into \emph{retrospective} and \emph{anticipatory} approaches. Retrospective methods assign reward after experience has occurred, using prediction error \citep{Pathak2017Curiosity}, state novelty \citep{burda2018exploration}, episodic novelty \citep{badia2020never}, or representation surprise \citep{Raileanu2020RIDE}. While simple and broadly compatible with model-free RL, they are vulnerable to the white-noise problem (attraction to aleatoric uncertainty) and detachment \citep{burda2018exploration, ecoffet2019go}. Anticipatory methods instead steer agents toward potentially novel states using short-horizon predictions of epistemic uncertainty \citep{shyam2019model, sekar2020planning, chua2018deep}. \rev{It can be argued that retrospective methods look for long term novelty as they attempt to influence the underlying agent behaviour gradually while anticipatory methods often react quickly to emerging novelty, making them a good method to seek short term novelty. We view our method as an anticipatory component that is orthogonal to and composable with retrospective / intrinsic reward methods.}
\subsection{Planning}
Planning in RL ranges from tree search to trajectory optimization. Monte Carlo Tree Search (MCTS) has proved effective in games \citep{mcts, alphago, alphazero} but good performance in these methods assumes (near) full observability and discrete action spaces. \rev{Although recent work (\cite{hubert2021learning, antonoglou2021planning}) has extended MCTS to stochastic dynamics and continuous actions, these works rely on models being highly accurate and tractable and are not able to be applied to stochastic, partially observed environments.} Path-integral / MPPI-style control samples and reweights trajectories under learned dynamics \citep{pspic, mppi}; TD-MPC and TD-MPC2 pair this with TD learning for vision-based control \citep{tdmpc, tdmpc2}, but the planner’s actions can drift from the policy network, risking distribution shift and value overestimation. Closer to our setting, Look Before You Leap prefers low-entropy, high-reward states, which can suppress exploration early in training \citep{lbyl}; MaxEnt-Dreamer \rev{maximizes a discounted entropy of the latent state-visitation distribution via an auxiliary density model and uses this only as a training-time regulariser on the actor, whereas our planner uses the world model’s own predictive entropy online to re-rank candidate trajectories, allowing it to react immediately to newly emerging high-uncertainty futures while still preferring rewarding ones} \citep{med}; RAIF optimizes posterior over prior uncertainty but this necessitates evaluating gains retrospectively, blind to emerging novelty \citep{raif}.
\subsection{Hierarchical Policies}
Hierarchical RL introduces temporal abstraction via high- and low-level controllers. Option-Critic learns options and termination conditions end-to-end \citep{oc}, while HiPPO runs PPO at two temporal scales \citep{hippo}. These methods improve long-horizon credit assignment, but fixed intervals or frequent replanning can limit adaptability; excessive termination also reduces effective commitment. Our planner differs by explicitly learning when to replan versus commit, driven by signals computed from imagined rollouts.
\section{Preliminaries}
\label{sec:preliminaries}

\rev{We consider a partially observable Markov decision process (POMDP)
\(P = \langle \mathcal{S}, \mathcal{A}, p, r, \mathcal{X}, Z, \gamma, \rho_0 \rangle\),
where \(\mathcal{S}\) is the state space, \(\mathcal{A}\) the action space, \(p\) the transition kernel,
\(r\) the reward function, \(\mathcal{X}\) the observation space, \(Z\) the observation kernel,
\(\gamma \in (0, 1)\) the discount factor, and \(\rho_0\) the initial-state distribution.
The agent observes pixels \(x_t \sim Z(\cdot \mid s_t)\) and acts via a policy
\(\pi(a_t \mid x_{1:t}, a_{1:t-1})\). A full MDP/POMDP formalism and its connection to
Dreamer-style world models is given in Appendix~\ref{sec:appendix_mdp_formalism}.}

Dreamer~\citep{hafner2019dream, hafner2020mastering, hafner2023mastering} learns a compact latent
dynamics model and performs policy optimization entirely in latent space. In this work, we use
DreamerV3 as it is the most recent and advanced formulation of the Dreamer series of models. At its
core, Dreamer relies on an RSSM that factorizes the environment into deterministic recurrent states
and stochastic latent representations. The RSSM is described by the following formulations:
\begin{align*}
\text{Recurrent model:} \quad & h_t = f_\phi(h_{t-1}, \rz_{t-1}, \ra_{t-1}) \\
\text{Transition predictor (prior):} \quad & \hat{\rz}_t \sim p_\phi(\cdot \mid h_t) \\
\text{Representation model (posterior):} \quad & \rz_t \sim q_\phi(\cdot \mid h_t, \rx_t) \\
\text{Predictors (Image, Reward, Discount):} \quad & \hat{x}_t, \hat{r}_t, \hat{\gamma}_t \sim p_\phi(\cdot \mid h_t, \rz_t)
\end{align*}
Here, $\rx_t$ is the observation at time $t$, and $h_t$ is the deterministic recurrent state. At each
step, Dreamer generates a prior latent state $\hat{\rz}_t$ from the deterministic recurrent state
$h_t$ via $p_\phi(\cdot \mid h_t)$, and updates it into a posterior $q_\phi(\cdot \mid h_t, \rx_t)$
once the new observation $\rx_t$ has been received. The KL divergence between the prior and
posterior is minimized to train the model:
\begin{equation}
\label{eq:kl_loss}
\mathcal{L}_{\text{KL}} = \KL\Big(q_\phi(\rz_t \mid h_t, \rx_t) \,\Vert\, p_\phi(\cdot \mid h_t)\Big).
\end{equation}
Dreamer’s policy network is trained using imagined trajectories generated by the world model. This
ensures that policy training remains effectively on-policy. The buffer that the world model trains
from is populated by a naive $\epsilon$-greedy actor.

\section{Method}
\label{sec:method}
\subsection{Entropy}
\rev{We now connect Dreamer’s KL term to an information-gain objective and motivate entropy as a practical, model-aligned uncertainty signal for planning. DreamerV3 models latent states as factorised discrete variables. The latent state $\rz_{t}$ is modelled by DreamerV3’s discrete RSSM, but the option to use a continuous latent space is also available. Training the model involves minimizing a KL divergence loss between the prior and posterior distributions of the latents (\eqref{eq:kl_loss}); classically, the same KL divergence can be viewed as an information gain (IG) term: for a target variable \(Y\) and an input \(X\), Quinlan’s information gain can be written as the reduction in entropy of \(Y\) after observing \(X\),}
\begin{equation}
\label{eq:ig_quinlan}
\rev{\text{IG}(Y; X)
\;=\;
H(Y) - H(Y \mid X)
\;=\;
\sum_{x} p(x)\,
\KL\bigl(p(Y \mid X = x)\,\Vert\,p(Y)\bigr)}
\end{equation}
\rev{highlighting that information gain is a form of mutual information, expressible as an expected KL divergence~\citep{quinlan1986induction}. In our setting, \(Y\) corresponds to the latent state \(\rz_t\) and \(X\) to the new observation \(\rx_t\). The training objective \(\KL\!\bigl(q_\phi(\rz_t \mid h_t, \rx_t)\,\Vert\,p_\phi(\rz_t \mid h_t)\bigr)\) is therefore the information gained about \(\rz_t\) from incorporating \(\rx_t\), up to averaging over time. At planning time, however, future observations \(\rx_t\) are not yet available, so the posterior \(q_\phi(\rz_t \mid h_t, \rx_t)\) cannot be evaluated for candidate action sequences. Any anticipatory uncertainty signal must therefore be computable from the prior alone.}

\rev{A simple proxy for future information gain is to prefer states whose prior \(p_\phi(\rz_t \mid h_t)\) has high entropy. Intuitively, if the prior over \(\rz_t\) is already low entropy, then, on average over possible observations, little uncertainty can be removed; conversely, a high entropy prior admits the possibility of large reductions in uncertainty. This motivates using the prior entropy as an intrinsic objective for the meta-planner. Thus, our objective becomes maximising prior entropy:}
\begin{equation}
\label{eq:prior_entropy_objective}
\rev{\mathcal{J}_t
\;=\;
\max \; H\bigl(p_\phi(\rz_t \mid h_t)\bigr)}
\end{equation}

\rev{The standard entropy functional $H[p] = \mathbb{E}_{X \sim p}[-\log p(X)]$ admits parallel definitions for discrete (Shannon) and continuous random variables, obtained by replacing the sum with an integral~\citep{CoverThomas2006,Shannon1948,MarshContinuousEntropy}. An advantage therefore of using entropy as an uncertainty measure is that it applies in a unified way across both categorical and Gaussian latent spaces without requiring changes to the model parameterisation.}

\rev{Thus, by selecting states with high prior entropy, the planner preferentially visits regions where the model’s beliefs are uncertain and observations are expected to concentrate the posterior distribution, increasing the KL divergence between the two. The planning objective therefore heuristically maximises the expected KL, while world-model training simultaneously minimises the same KL term, yielding a natural (albeit loose) minmax interaction between exploration and model fitting.}

There are two primary failure modes for this uncertainty-based exploration. The first arises in environments with high aleatoric uncertainty, where a state has multiple plausible successors due to \rev{inherent stochasticity}. In such cases, even a well-understood state \(s_t\) may give rise to a high-entropy predictive latent distribution simply because \rev{there are multiple legitimate outcomes. This occurs because the RSSM uses a unimodal prior family. During training, the prior \(p_\phi(\rz_{t+1} \mid h_t)\) must match the posterior via the KL term, but the posterior only captures the realised outcome, not the full set of possibilities. When the true next-latent distribution is effectively multi-modal,}
\begin{equation}
\rev{p^{\ast}(\rz_{t+1} \mid h_t) 
= \sum_{i=1}^{M} w_i \, p_i(\rz_{t+1} \mid h_t),
\qquad
w_i \ge 0,\;\;\sum_{i=1}^M w_i = 1}
\end{equation}
\rev{the RSSM is forced to approximate all modes with a single member of its unimodal family,}
\begin{equation}
\rev{\hat{p}_\phi(\rz_{t+1} \mid h_t) \in \mathcal{F}_{\text{uni}},
\qquad
H\bigl(\hat{p}_\phi(\rz_{t+1} \mid h_t)\bigr)
\;\gg\;
H\bigl(p_i(\rz_{t+1} \mid h_t)\bigr)
\;\;\text{for many } i}
\end{equation}
\rev{so that the learned prior entropy is artificially inflated. Here \(i\) indexes the distinct plausible successor states and the weights \(w_i\) describe their relative frequencies. These states are not inherently bad to be in: they are often bottleneck states that lead to many other states \citep{ecoffet2019go}, some rewarding and some not. Our agent is naturally encouraged to visit such bottlenecks, but because imagined trajectories are scored by both environment reward and entropy, the greedy actor will favour branches that reach reward-relevant regions, which mitigates pathological fixation on such states.}

The second failure mode arises in environments with latent transitions that require specific, rarely executed actions. \rev{In such cases, the ideal transition distribution may again be written as a mixture,}
\begin{equation}
\rev{p^{\ast}(\rz_{t+1} \mid h_t) 
= \sum_{i=1}^{M} w_i \, p_i(\rz_{t+1} \mid h_t),
\qquad
w_i \ge 0,\;\;\sum_{i=1}^M w_i = 1}
\end{equation}

\rev{but now the weight associated with the common transitions, denoted \( w_C \), satisfies
\( w_C \gg w_{\lnot C} \), where \( w_{\lnot C} \) is the total weight of all rare transitions. If
the agent has only encountered the high-probability modes, the learned model will be ignorant of the
rare outcomes and will instead estimate a prior dominated by the common component,}
\begin{equation}
\rev{\hat{p}_\phi(\rz_{t+1} \mid h_t)
\approx p_C(\rz_{t+1} \mid h_t)}
\end{equation}
\rev{so that}
\begin{equation}
\rev{H\bigl(\hat{p}_\phi(\rz_{t+1} \mid h_t)\bigr)
\;\approx\;
H\bigl(p_C(\rz_{t+1} \mid h_t)\bigr)
\;<\;
H\bigl(p^{\ast}(\rz_{t+1} \mid h_t)\bigr)}
\end{equation}
\rev{In these cases, a state's uncertainty may be chronically underestimated and subsequently underexplored. Addressing this likely requires mode-seeking mechanisms (option discovery, social learning, teacher-student learning). While this is outside the scope of the present method, it is possible to amplify this method with future work.}

\subsection{Reactive hierarchical planner}
\label{sec:planner}
At each environment step we generate $N$ short-horizon imagined rollouts (we use \rev{$N{=}256$) starting from the current latent state using the greedy actor and the world model. We then select the rollout trajectory $\tau_\star$ whose latent prior has the highest cumulative entropy \emph{and} highest cumulative reward ($\hat{r}_t$ as predicted by dreamer's internal reward model):
\begin{equation}
\label{eq:plan_selection}
\tau_\star 
= \arg\max_{\tau \in \{\tau^{(n)}\}_{n=1}^N}
\sum_{t' = t}^{t + H}
\Big(
    \lambda_r \hat{r}_{t'} \;+\; \lambda_H H\bigl(p_\phi(\rz_{t'} \mid h_{t'})\bigr)
\Big).
\end{equation}
where both $\lambda_r$ and $\lambda_H$ set the relative weighting of entropy and reward and $H$ is the planning horizon: we use 15 for $H$ as it is the length to which dreamer's world model is trained by default. We then execute the selected trajectory's set of actions (a plan) unless the meta-policy decides to cancel the current plan and generate a new one. We find in practice that our method is not overly sensitive to $\lambda_r$ and $\lambda_H$, we do however test this by ablating just the reward and then just the entropy.}

\rev{We use a meta planner (a light PPO head) to control when to replan.} The meta planner outputs a categorical over $p_t\!\in\!\{0,0.25,0.5,0.75,1\}$ which we squashed as $p_t^2$ to decide replanning: draw $u_t\!\sim\!\mathcal U(0,1)$ and replan if $u_t < p_t^2$. We find in practice that excessive replanning is a problem that frequently plagues such planners, a finding echoed across the hierarchical RL literature~\citep{switch1,switch2,switch3} which is why we change the enaction condition from $\ru_t < p_t$ to $\ru_t < p_t^2$, discouraging excessive replanning without explicitly punishing replanning or rewarding commitment. \rev{Empirically, the learned meta-policy settles into short but non-myopic commitments, with plan lengths of ~2–3 steps on average and occasional longer bursts (see Appendix \ref{sec:appendix_planner_metrics}, Planner Metrics).}

Importantly, planning decisions are re-evaluated at every environment step, allowing for flexible replanning without commitment if needed. Pseudocode for our planning algorithm is given in Appendix \ref{sec:appendix_B}. As input to the PPO policy, we provide the encoder embedding; the RSSM feature vector; the current step number normalized by the episode time limit; the greedy action proposed by the actor; the position within the current plan (normalized); a binary in-plan flag; and the “final” RSSM feature that is predicted to be observed if the current plan is followed to the end.

We maintain replay buffers of meta-transitions with fields: PPO observation, PPO action, PPO sample log prob, implemented flag (whether a replan signal was sent), per-step entropy, next base reward, and done. \rev{The PPO meta-planner is trained on sequence-level returns, aggregating the reward and entropy terms over a planning horizon of length $L$:
\begin{equation}
\label{eq:meta_reward}
r_{t}^{\text{meta}}
\;=\;
\tfrac{L}{2}
\sum_{t' = t}^{t + L}
\Big(
    r_{t'}^{\text{base}} \;+\; H\bigl(p_\phi(\rz_{t'} \mid h_{t'})\bigr)
\Big).
\end{equation}}
We then compute advantages with GAE$(\gamma,\lambda)$ and optimize a clipped PPO objective (separate actor/critic) with a naive entropy bonus, using Adam for both policy and value heads. The PPO head trains on all collected transitions. To encourage early behavioral diversity we use He initialization and bias the PPO head’s initial logits toward intermediate $p_t$ values (0.25, 0.5, 0.75).

\rev{In general, the choice of $L$ controls the temporal scope of the meta-planner’s objective: long,
sparse-reward or highly delayed-reward tasks benefit from larger $L$, which allows the planner to
align its decisions with far-reaching consequences, whereas short-horizon, locally reactive tasks
favour smaller $L$ to enable fast adaptation. In our experiments, we set $L{=}32$ for complex,
long-horizon environments such as Crafter, $L{=}8$ for navigation tasks (MiniWorld Maze), and
$L{=}2$ for short-horizon continuous control tasks (DMC-Vision). This choice lets the reward
calculation match the intended commitment level required by each environment.}
\section{Experiments}
\label{sec:experiments}
We evaluate across three regimes that stress different aspects of decision making: procedurally generated 3D mazes in MiniWorld \cite{MinigridMiniWorld23} for long-horizon navigation under partial observability and sparse reward, Crafter (\cite{crafter}) for survival-style open-world play with diverse subgoals, and vision-based control tasks from the DeepMind Control suite (\cite{dmc}) for closed-loop continuous control. We report means with variability over 5 seeds for MiniWorld [0, 409, 412, 643, 996] and DMC [0, 413, 604, 765, 891], and \rev{5} seeds for Crafter [0, 11, \rev{413, 891,} 920]. Because MiniWorld and Crafter are procedurally generated, training and test distributions coincide; MuJoCo-based DMC tasks have no train/test split under our setup; \rev{we therefore report training curves only.}

We compare with Plan2Explore (\citet{sekar2020planning}) as a baseline; it supplies an anticipatory exploration baseline that scores novelty via ensemble disagreement, giving a contrast between disagreement-based uncertainty and our inference-time entropy signal from a single world model. We also add PPO on MiniWorld as a model-free reference \citep{schulman2017ppo}. For Crafter and DMC, competitive pixel-based model-free methods (augmented SAC/DrQ-style agents) are known to be sensitive to implementation and tuning \citep{engstrom2020implementation,henderson2018matters}; to avoid confounding factors and keep compute comparable, we omit them here \citep{kostrikov2020image,yarats2021drqv2,laskin2020rad,srinivas2020curl}.
Dreamer is a widely adopted and strong pixel-based MBRL agent that already surpasses earlier model-based and many model-free methods in visual control \citep{hafner2019dream, hafner2020mastering, hafner2023mastering}.

We train MiniWorld mazes for 350{,}000 environment steps (approximate convergence for Dreamer under our setup). On DMC and Crafter we use a fixed 24-hour wall-clock budget per method to ensure compute parity. Metrics are task-appropriate: episode time to completion for MiniWorld (shorter is better) and undiscounted training return for DMC and Crafter. \rev{All curves use a rolling mean (window 10\%) with shaded $\pm$1 standard error of the mean (SEM) across seeds.}

\subsection{Maze Explorer}
We use a 3D maze environment adapted from MiniWorld \citep{MinigridMiniWorld23}, where each episode presents a new random layout. The agent receives RGB image observations and performs continuous actions to locate three goal boxes. We introduce a \textit{porosity} parameter that controls wall density to vary exploration difficulty. Observations are augmented with a binary spatial map that encodes visited regions and current orientation, providing a simple episodic memory. The reward function combines exploration, proximity, and goal rewards to encourage both coverage and task success. Full environment details are in Appendix~\ref{sec:appendix_C}.

\subsubsection{Task Difficulty}

Varying the porosity parameter controls maze difficulty (low porosity leads to difficult mazes). Visual examples of mazes at different porosity levels are provided in Appendix \ref{sec:appendix_F}. Figure~\ref{fig:train_length_plot} shows that our method maintains low episode lengths even in denser mazes, outperforming both Dreamer and PPO. PPO underperforms across all settings, likely due to its lack of memory and long-horizon reasoning. Base Dreamer’s performance degrades under low porosity, where rewards are harder to reach, while our method maintains faster episode finishes across porosities in our runs.

Under the most difficult condition (porosity = 0), where only a single path exists between the agent and three goals, both our method and Dreamer perform worse\rev{, as shown in Fig.~\ref{fig:train_length_plot}}. However, our approach still outperforms Dreamer, achieving 20\% shorter episode lengths on average, albeit with higher variance due to the increased exploration burden. Plan2Explore underperforms here, which we hypothesize stems from frequent replanning without commitment; ensemble disagreement identifies novelty but does not enforce trajectory-level persistence.

\begin{figure}[tb]
    \centering
    \includegraphics[width=\linewidth]{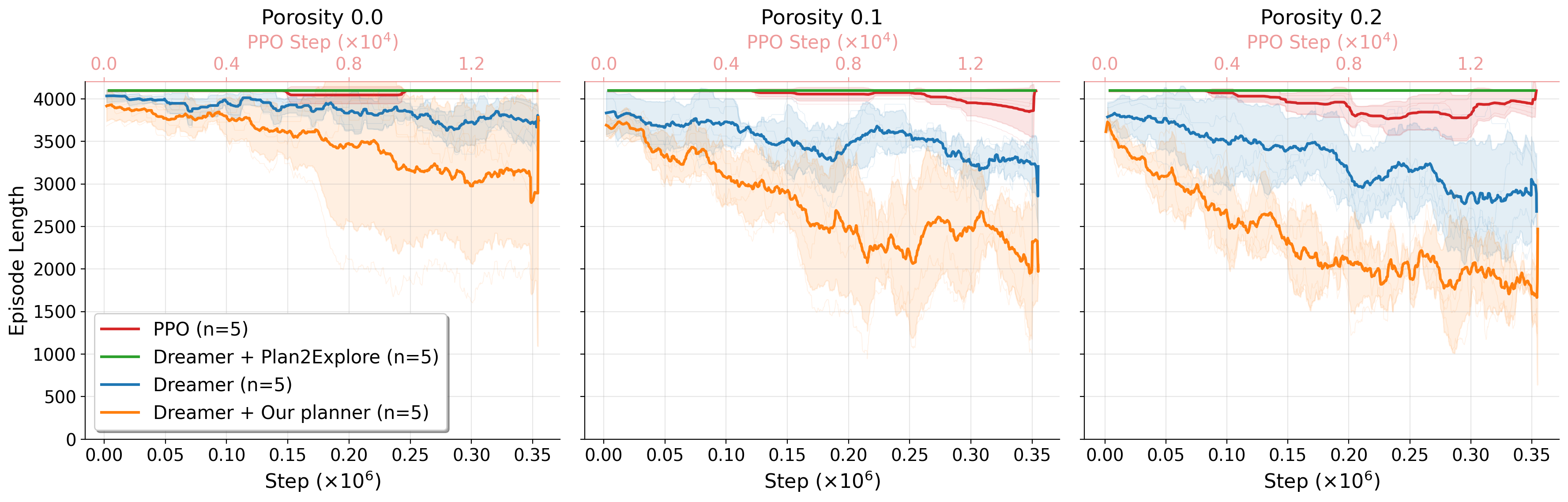}
    \caption{\rev{Episode lengths across different porosity levels. Lower porosity increases maze difficulty.}}
    \label{fig:train_length_plot}
\end{figure}

\subsection{Vision-based control (DMC-Vision)}
\label{sec:dmcvision}

We evaluate six pixel-control tasks from the DeepMind Control Suite: \texttt{cartpole\_swingup}, \texttt{walker\_walk}, \texttt{cheetah\_run}, \texttt{reacher\_hard}, \texttt{acrobot\_swingup}, and \texttt{hopper\_hop}. Rather than sweeping the entire benchmark, we select a compact set that spans complementary regimes. Agents observe $64{\times}64$ RGB frames and act in continuous spaces. For each task we train (i) a base Dreamer agent without planning, (ii) our commit-aware planning variant, and (iii) Plan2Explore, all under identical step budgets (in practice they also take up similar amounts of time, more detail is given in the timing analysis section); curves report \rev{a rolling mean (window 10\%) with shaded $\pm$2 standard error of the mean (SEM) across seeds. Unless otherwise stated, we use a plan horizon of $H{=}16$, a PPO reward length $L{=}2$, sample $N{=}256$ actor-guided candidate rollouts per decision.} To isolate reasoned exploration, we use a single environment instance (no vectorization), since heavy parallelism can introduce ``free'' random exploration. For the same reason, we omit sparse-reward variants (e.g., \texttt{cartpole\_swingup\_sparse}), where neither base Dreamer nor our planner is expected to reliably discover narrow reward regions within the given budget.

\rev{A small sensitivity sweep over candidate count and meta horizon on four of the six tasks (cartpole\_swingup, walker\_walk, hopper\_hop, reacher\_hard) indicates that gains are stable across a reasonable range of values (Appendix~\ref{sec:appendix_ablation_sensitivity}). We restrict these sweeps to a representative subset to control compute: acrobot\_swingup is another underactuated swing-up control task whose planner behaviour closely mirrors cartpole\_swingup, and cheetah\_run is a high-speed locomotion task already well covered by the hopper\_hop and walker\_walk sensitivity profiles, so additional sweeps on these two tasks would be computationally costly without providing qualitatively new insights.}

\begin{figure*}[t]
  \centering
  \captionsetup{font=small}
  \begin{subfigure}[t]{0.32\linewidth}
    \centering
    \includegraphics[width=\linewidth]{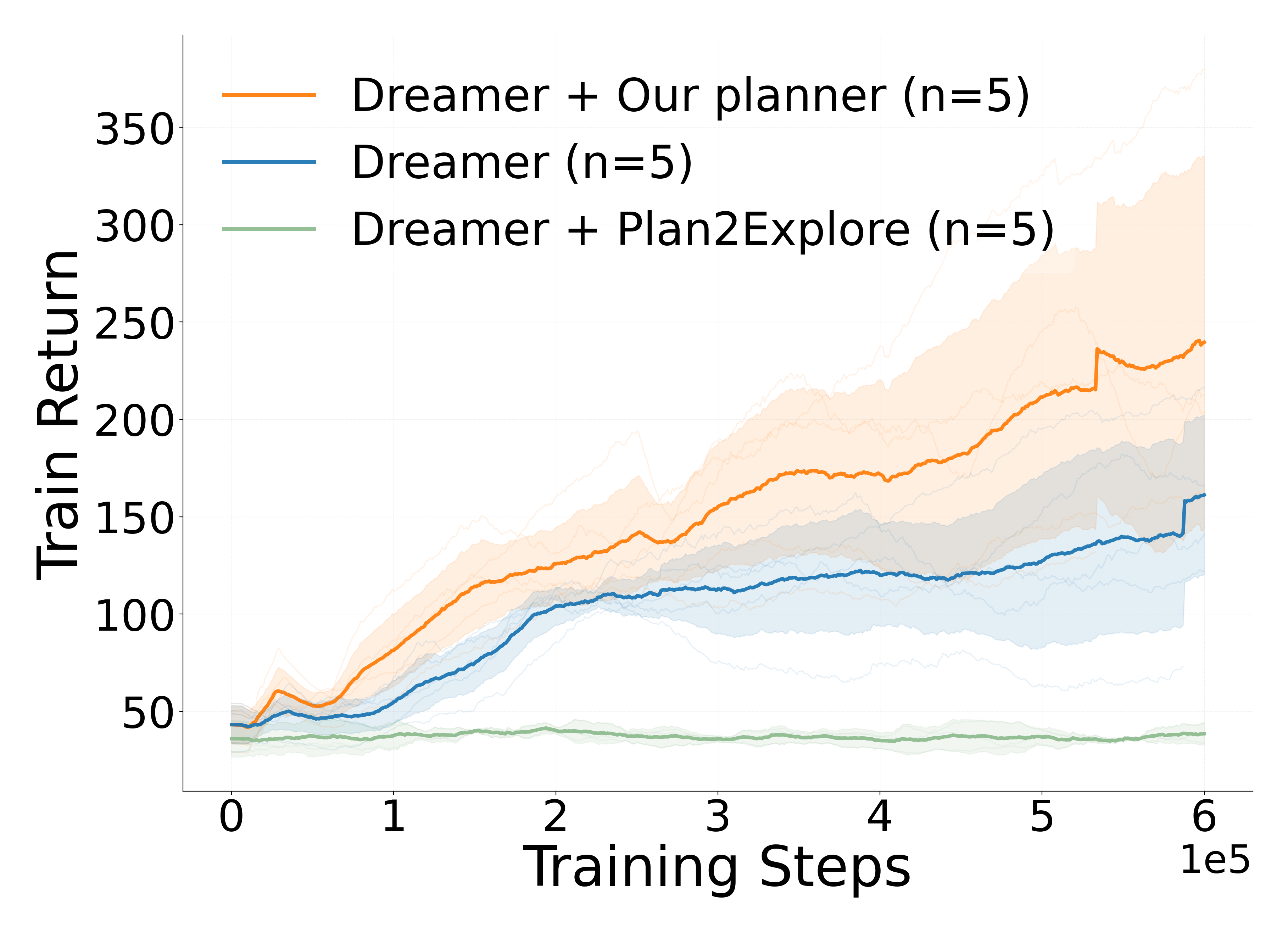}
    \caption{\rev{walker\_walk: planning maintains a widening lead over baseline.}}
    \label{fig:walker_walk}
  \end{subfigure}\hfill
  \begin{subfigure}[t]{0.32\linewidth}
    \centering
    \includegraphics[width=\linewidth]{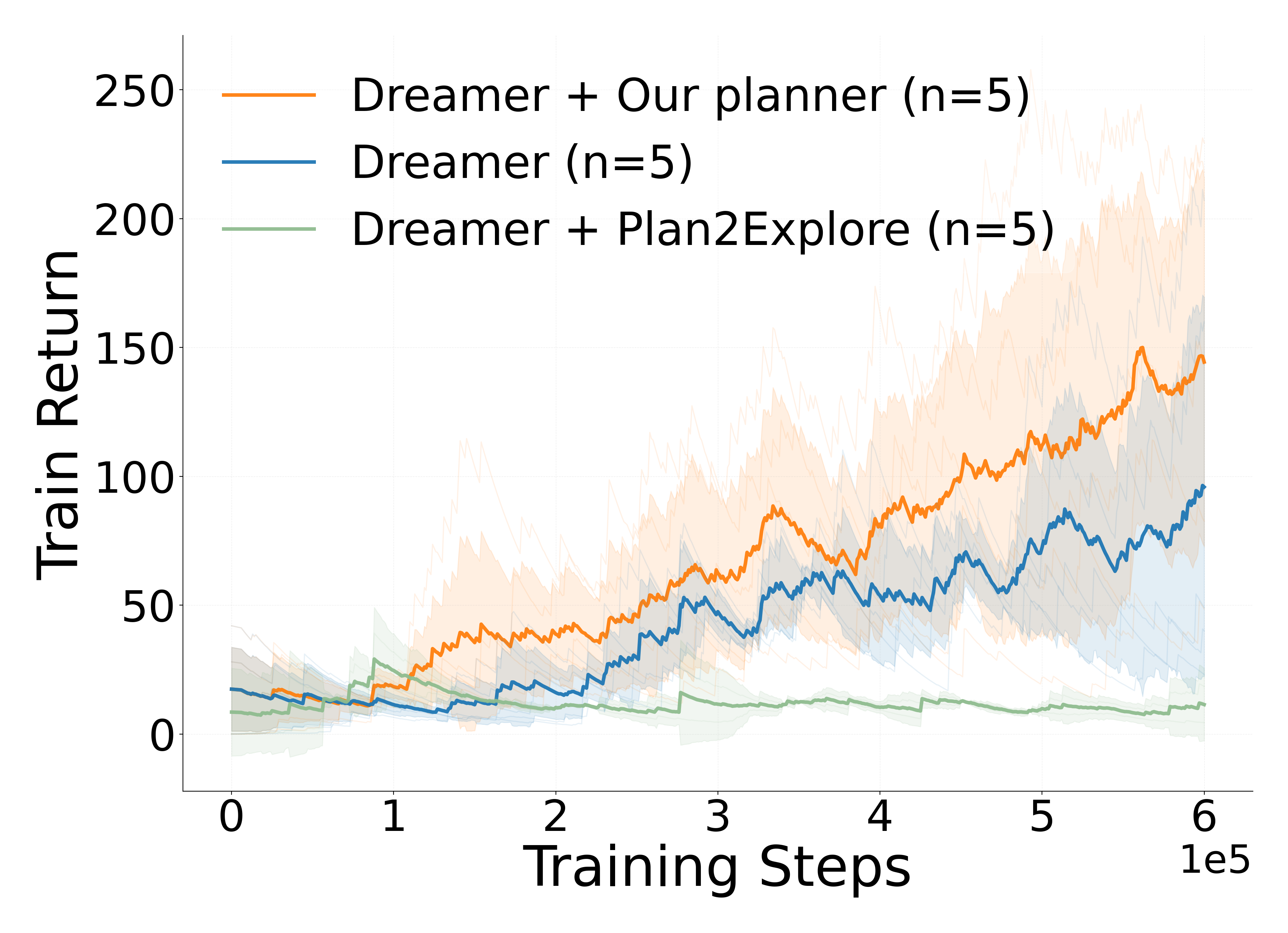}
    \caption{\rev{reacher\_hard: clear gains from purposeful multi-step corrections.}}
    \label{fig:reacher_hard}
  \end{subfigure}\hfill
  \begin{subfigure}[t]{0.32\linewidth}
    \centering
    \includegraphics[width=\linewidth]{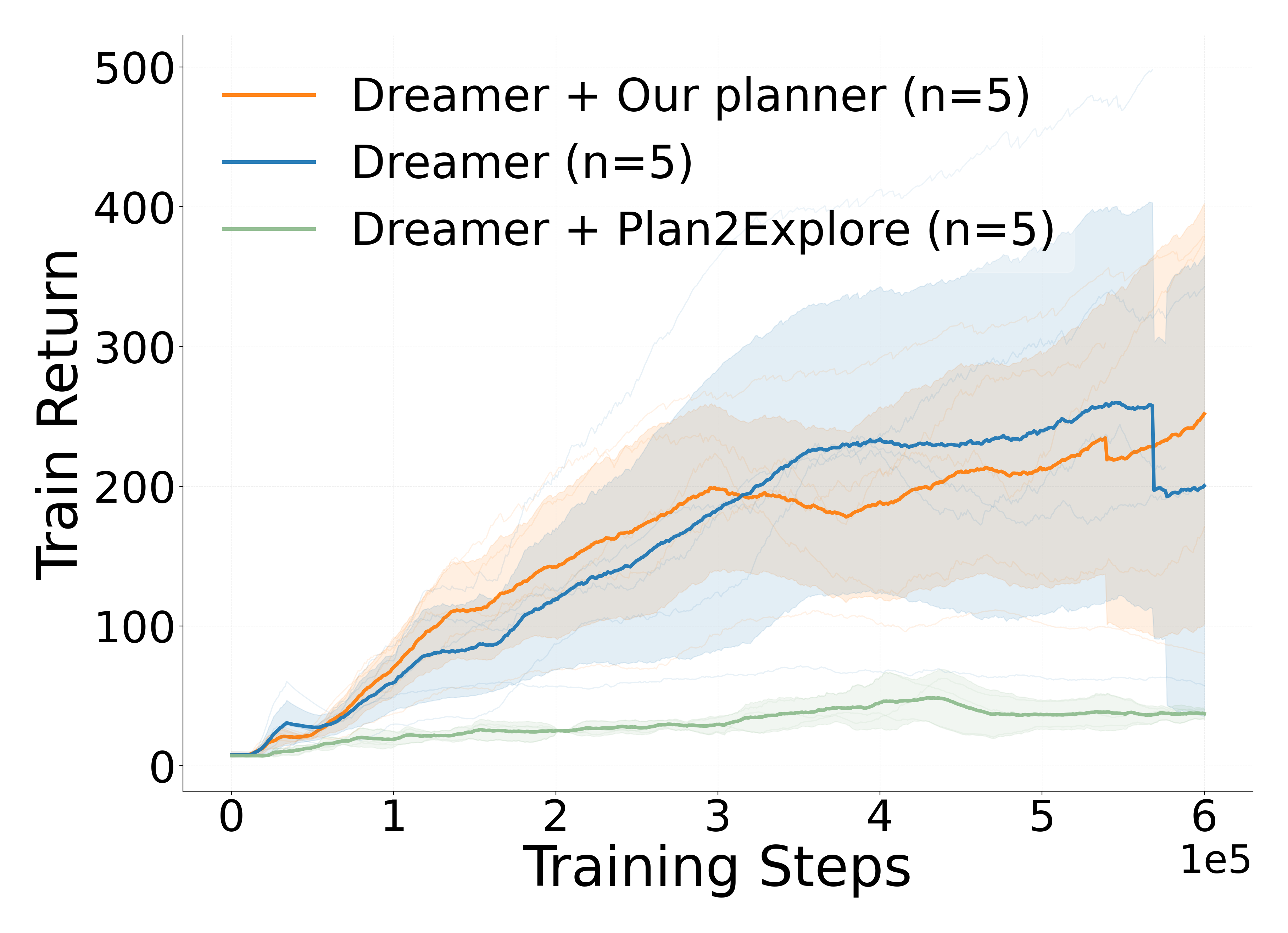}
    \caption{\rev{cheetah\_run: planning variant shows no improvement.}}
    \label{fig:cheetah_run}
  \end{subfigure}
  \begin{subfigure}[t]{0.32\linewidth}
    \centering
    \includegraphics[width=\linewidth]{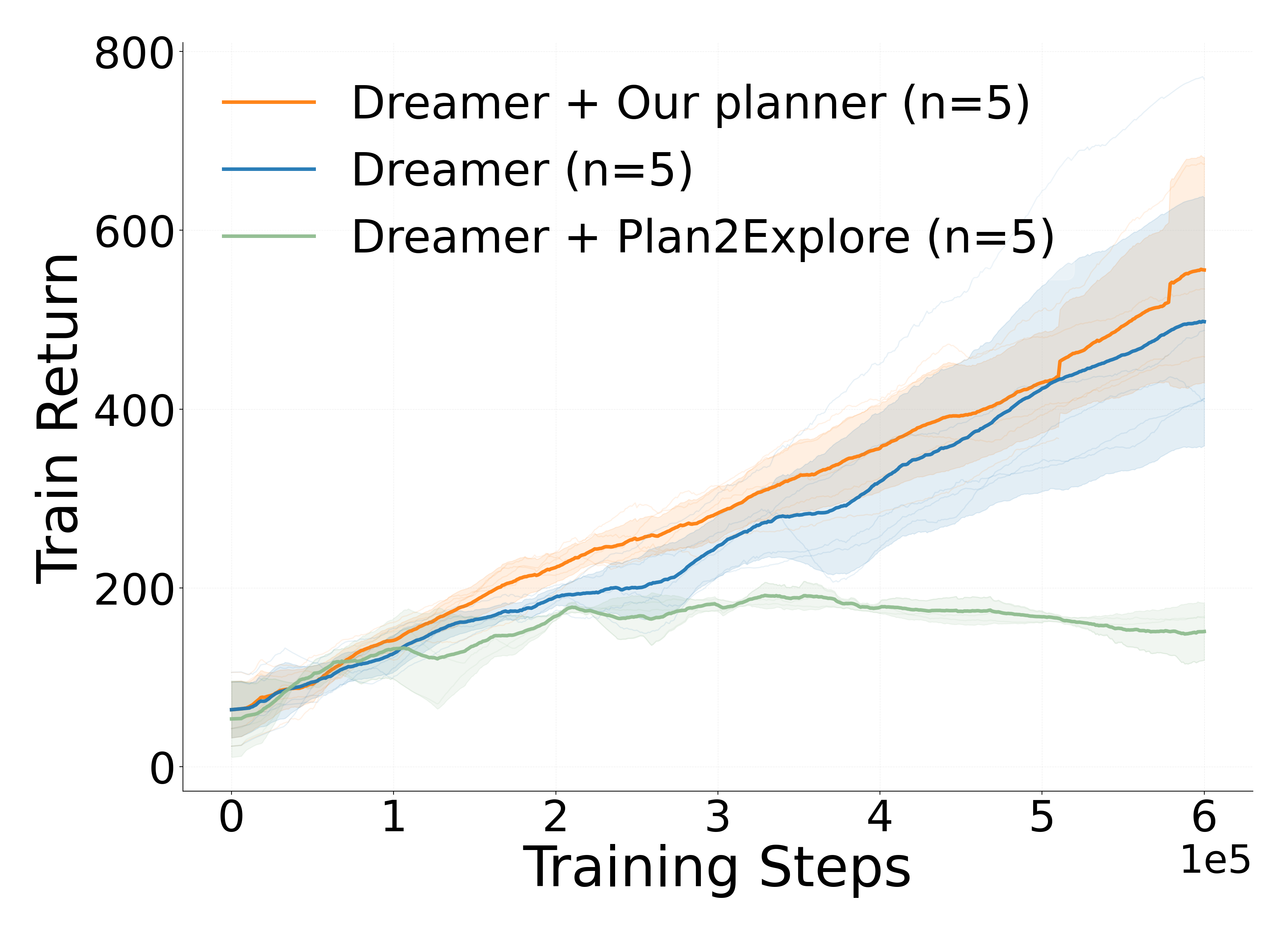}
    \caption{\rev{cartpole\_swingup: Not much change between planning and no planning variants.}}
    \label{fig:cartpole}
  \end{subfigure}\hfill
  \begin{subfigure}[t]{0.32\linewidth}
    \centering
    \includegraphics[width=\linewidth]{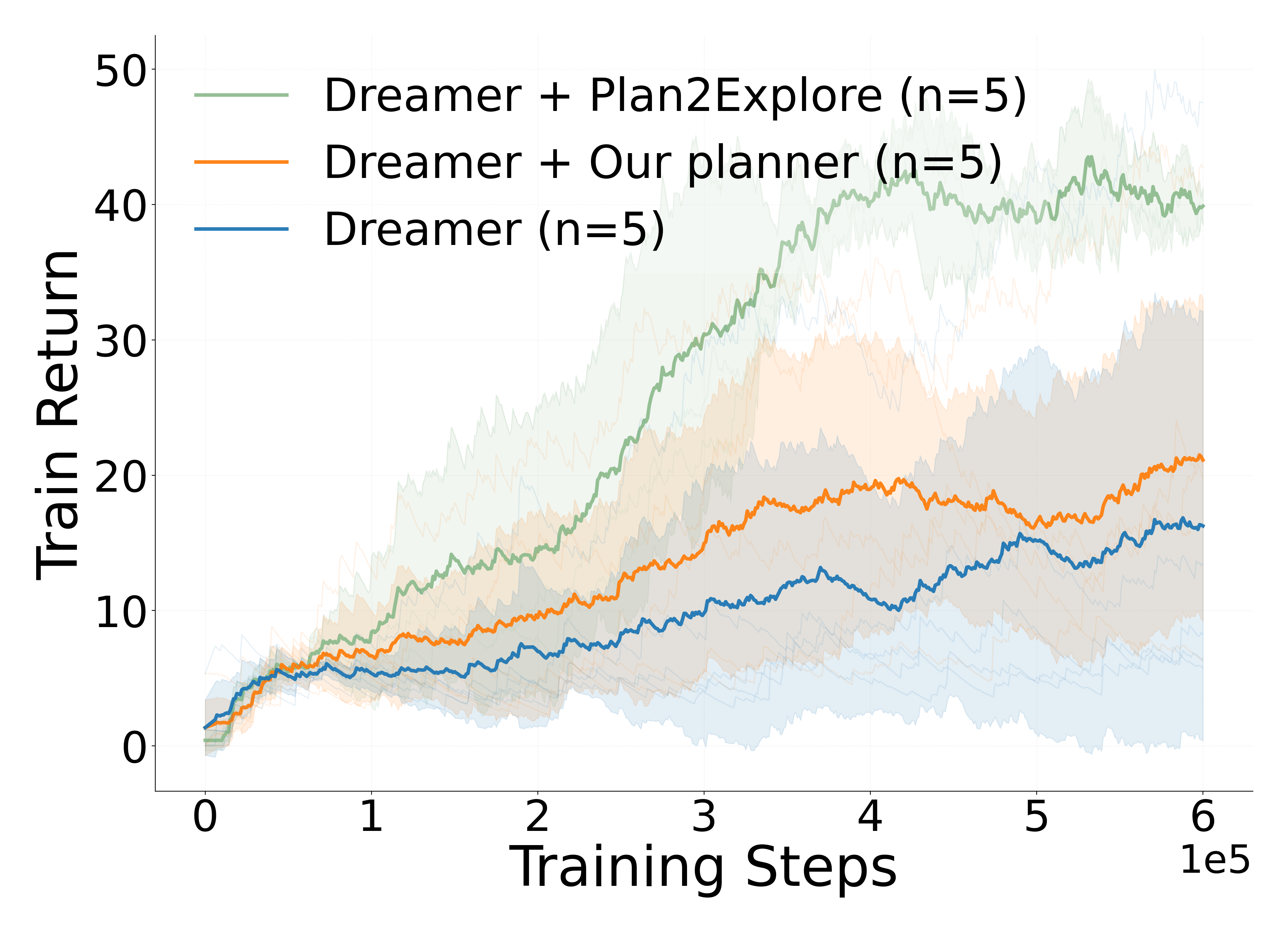}
    \caption{\rev{acrobot\_swingup: planning variant is on par or slightly better, but plan2explore greatly improves upon both here.}}
    \label{fig:acrobot}
  \end{subfigure}\hfill
  \begin{subfigure}[t]{0.32\linewidth}
    \centering
    \includegraphics[width=\linewidth]{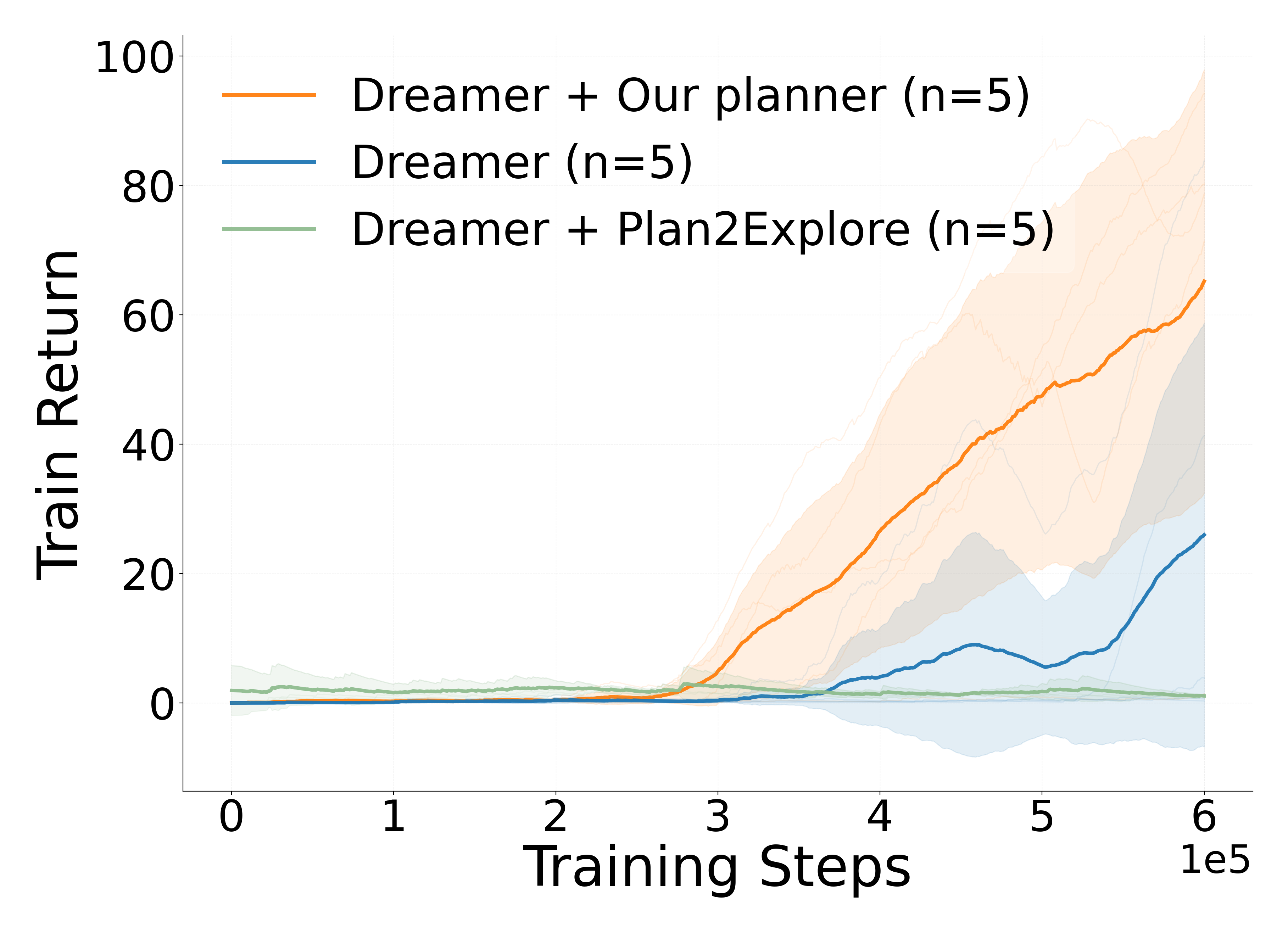}
    \caption{\rev{hopper\_hop: Our method starts finding the control sequences that will make the hopper hop much earlier than the no planning variant.}}
    \label{fig:hopper_hop}
  \end{subfigure}

  \caption{\rev{DMC-Vision learning curves (return vs.\ environment steps) for no-plan, the planning variant, and Plan2Explore. Shaded bands: $\pm$2 standard error of the mean (SEM) across seeds.}}
  \label{fig:dmc_grid_2x3}
\end{figure*}

Across \rev{three} of six tasks, the planning variant achieves higher sample efficiency and final return on average. Gains are largest in contact-rich or higher-variance control (Figures~\ref{fig:walker_walk}, \ref{fig:hopper_hop}), where short commitments reduce dithering and stabilize control under pixel noise while collecting informative trajectories. On smoother, lower-variance dynamics (Figures~\ref{fig:acrobot}, \ref{fig:cartpole}), improvements are smaller but positive on average. Despite \texttt{hopper\_hop} often benefiting from increased parallelism or longer training, our method discovers effective hopping sequences earlier than no-plan (Figure~\ref{fig:hopper_hop}). Plan2Explore underperforms on most tasks here but is strong on \texttt{acrobot\_swingup}, suggesting that disagreement-based novelty aligns with that underactuated swing-up; by contrast, our approach \rev{does not detract performance from any task but can make significant improvements on some of the more complex tasks that necessitate sequences of actions.}
\subsection{Open-ended survival (Crafter)}
\label{sec:crafter}
Crafter stresses long-horizon exploration and routine formation. \rev{Final episode rewards are given for the number of unique achievements collected, limited to 22. Small penalties and bonuses are given for health changes~\citep{crafter}.} We train for 300k environment steps (approximately 24 hours on a GeForce RTX 5090 GPU) and report mean returns with variability over 5 seeds. We run a single environment here as well (no parallel rollouts), which is the default for Crafter. \rev{Unless otherwise stated, the planning variant in Crafter is trained with the entropy-only meta objective (and without base reward).}

\begin{figure}[tb]
    \centering
    \includegraphics[width=0.6\linewidth]{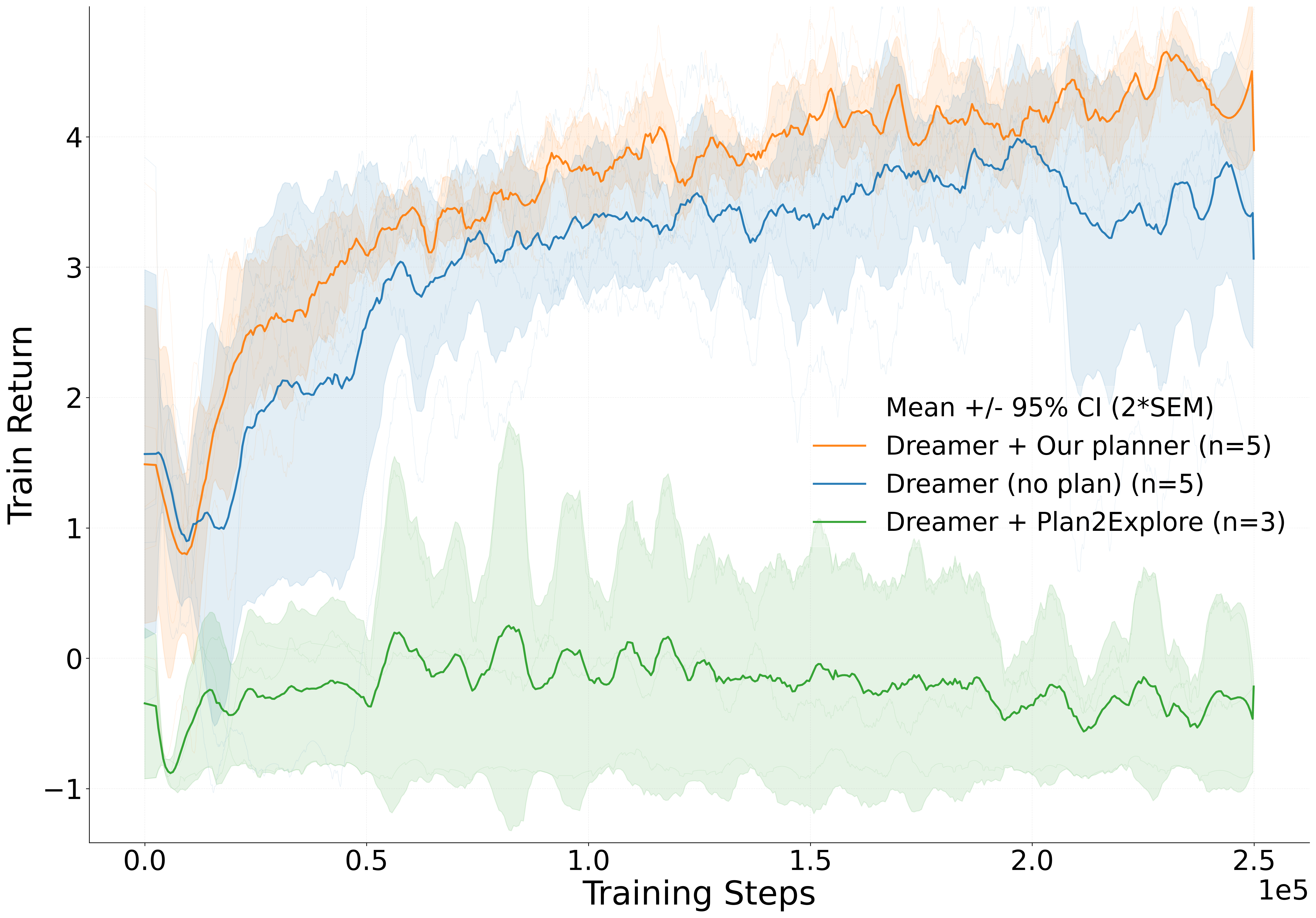}
    \caption{\rev{Episode returns during Crafter training.}}
    \label{fig:crafter_perf}
\end{figure}

Overall, the planning variant is about 20\% higher in average return and reaches comparable thresholds in roughly 50\% of the steps that base Dreamer takes (Fig.~\ref{fig:crafter_perf}). \rev{Because Crafter’s reward is dominated by exploratory achievements like crafting tools, placing structures, or engaging enemies, Dreamer’s value and policy networks already internalise a form of long-horizon, reward-driven curiosity. Our planner augments this implicit curiosity with an anticipatory entropy-based meta objective, so this comparison can also be read as an ablation between a curiosity-driven agent with and without anticipatory planning. The achievement breakdown in Appendix~\ref{app:crafter_achievements} shows that the planner-equipped variant acquires repeatable, reward-dense survival routines earlier and more consistently, while Plan2Explore underperforms on Crafter, consistent with the task rewarding sustained multi-step routines rather than one-step novelty chasing.}
\subsection{Ablation Study}
To isolate the contributions of individual components, we compare:
\begin{description}
    \item[Base Dreamer:] The standard agent without planning.
    \item[MPC Style:] To evaluate the value of our underlying planner, we modify the meta planner to replan at every step, mimicking MPC behaviour.
    \rev{\item[Meta-reward components:] We ablate the signal used to train the meta-policy by using (i) entropy-only, (ii) reward-only, and (iii) a 50/50 mixture of reward and entropy in $r_{\text{meta}}$.}
\end{description}

We run all MPC ablation variants for 90\% of the normal experiments' step count to account for the fact that MPC style experiments are slower to run than the full experiments.

\begin{figure}[tb]
    \centering
    \begin{subfigure}[t]{0.48\linewidth}
        \centering
        \includegraphics[width=\linewidth]{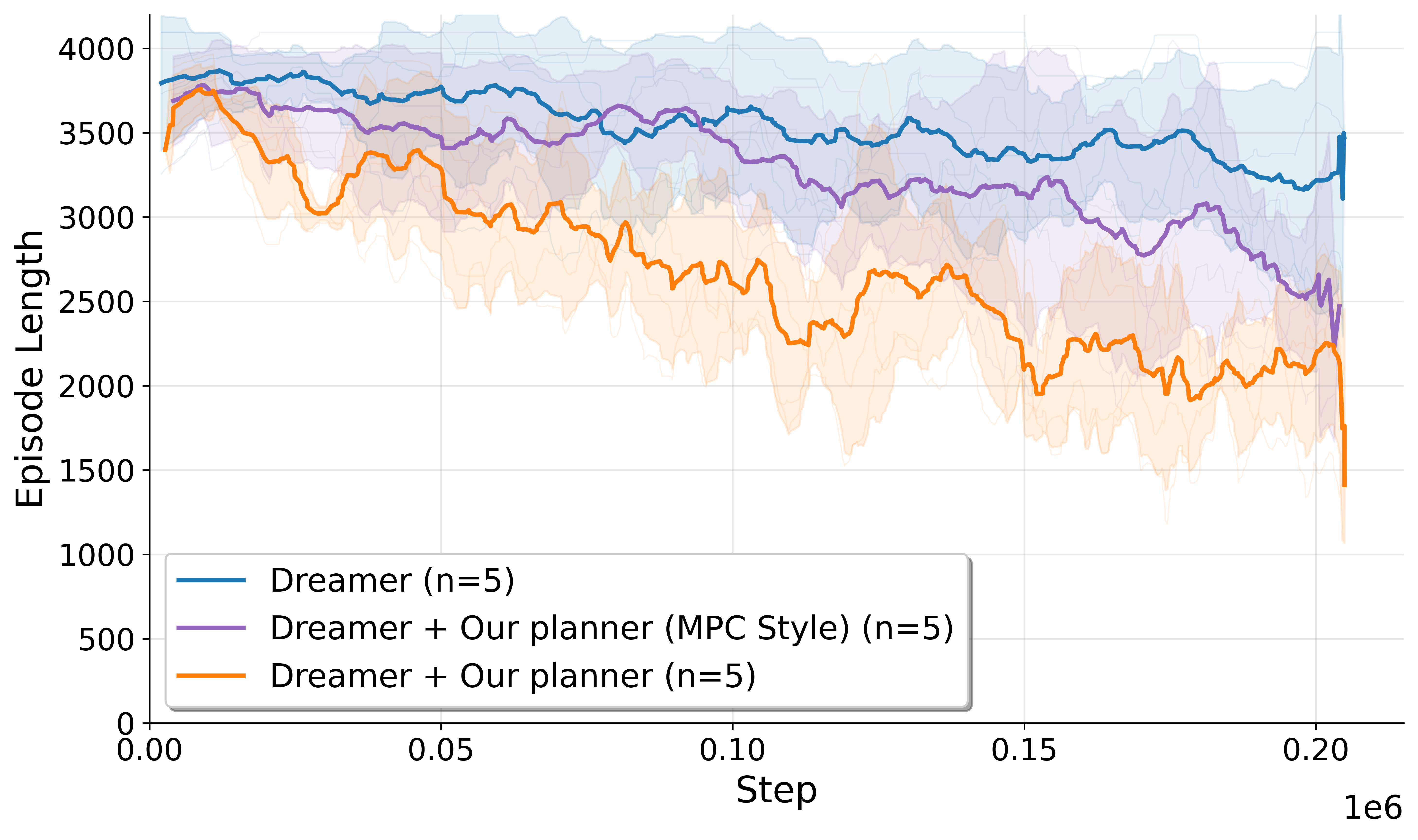}
        \caption{\rev{ObjectNav: Episode lengths during training for different ablations. 
        The full planner outperforms both the base Dreamer and MPC-style variants, 
        highlighting the benefit of plan commitment.}}
        \label{fig:train_length_ablate_objectnav}
    \end{subfigure}
    \hfill
    \begin{subfigure}[t]{0.48\linewidth}
        \centering
        \includegraphics[width=\linewidth]{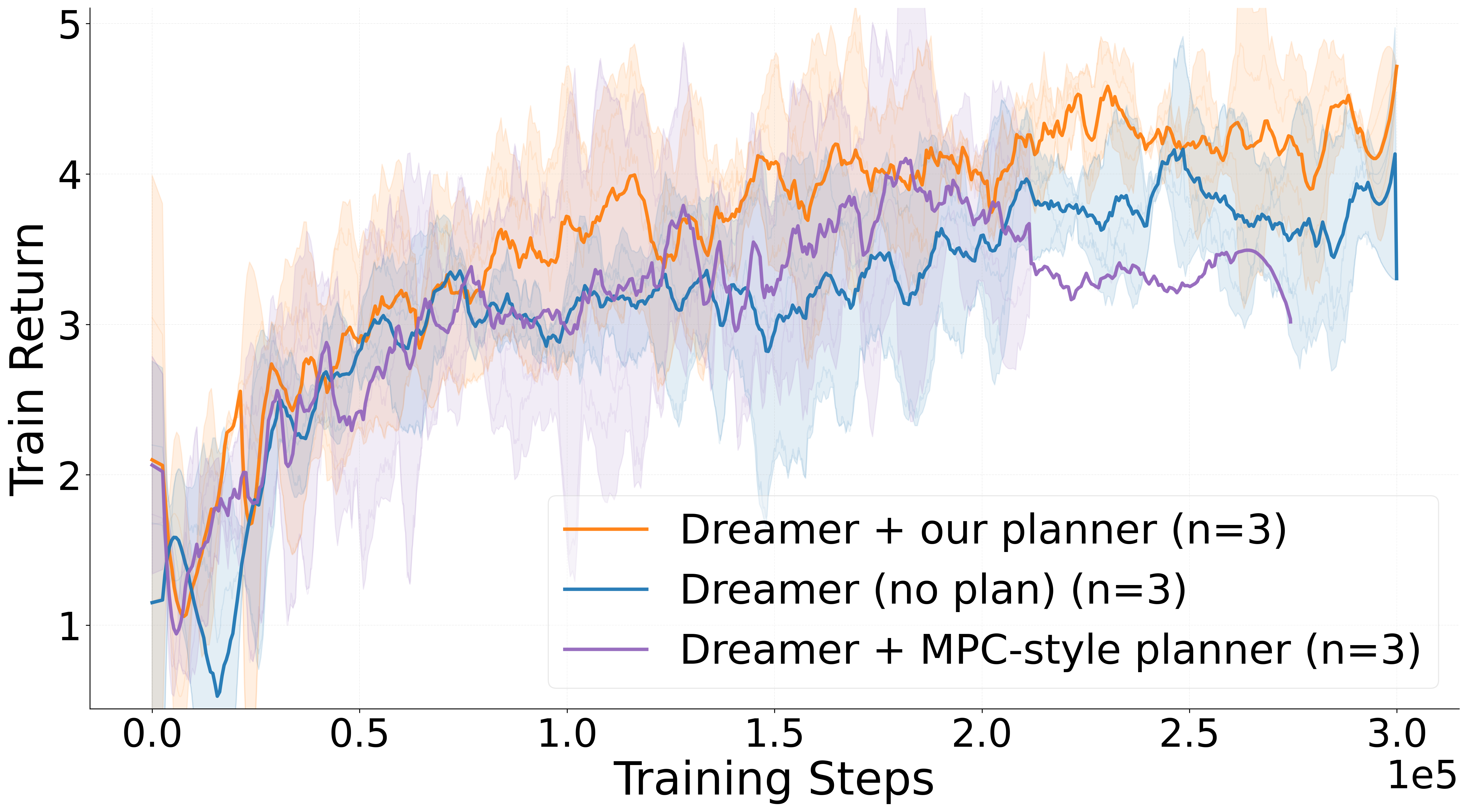}
        \caption{\rev{Crafter: Episode returns during training for different ablations. 
        The full planner has greater performance and takes less time than the MPC variant, signalling efficiency benefits of plan commitment.}}
        \label{fig:train_length_ablate_crafter}
    \end{subfigure}
    \caption{Comparison of episode lengths during training for ObjectNav (left) and Crafter (right) across different ablations.}
    \label{fig:combined_ablation_fig}
\end{figure}

\begin{figure}[tb]
    \centering
    \begin{subfigure}[t]{0.49\linewidth}
        \centering
        \includegraphics[width=\linewidth]{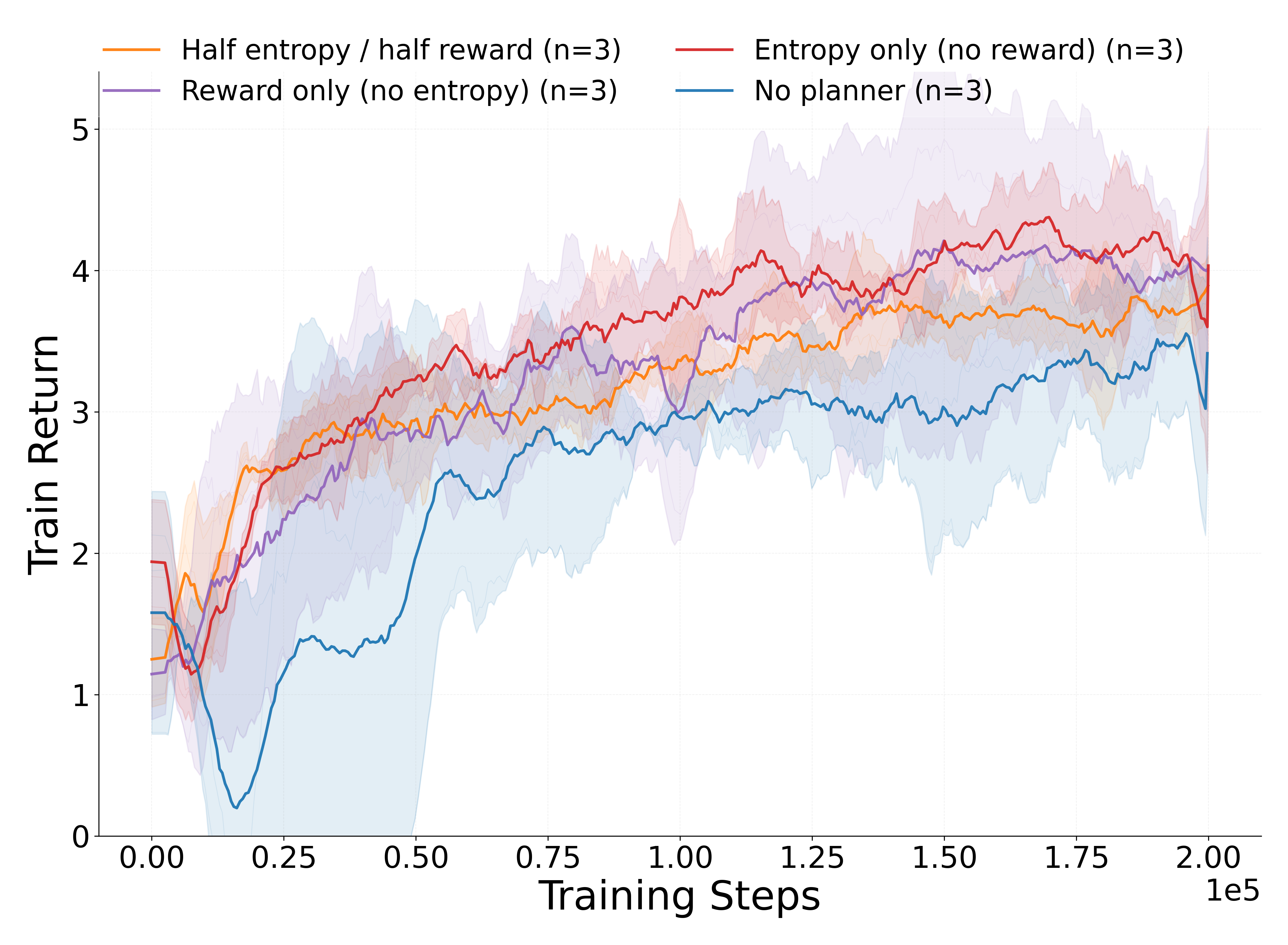}
        \caption{\rev{Crafter entropy/reward ablation. All three planner traces outperform the base dreamer variant}}
        \label{fig:crafter_entropy_ablate}
    \end{subfigure}
    \hfill
    \begin{subfigure}[t]{0.49\linewidth}
        \centering
        \includegraphics[width=\linewidth]{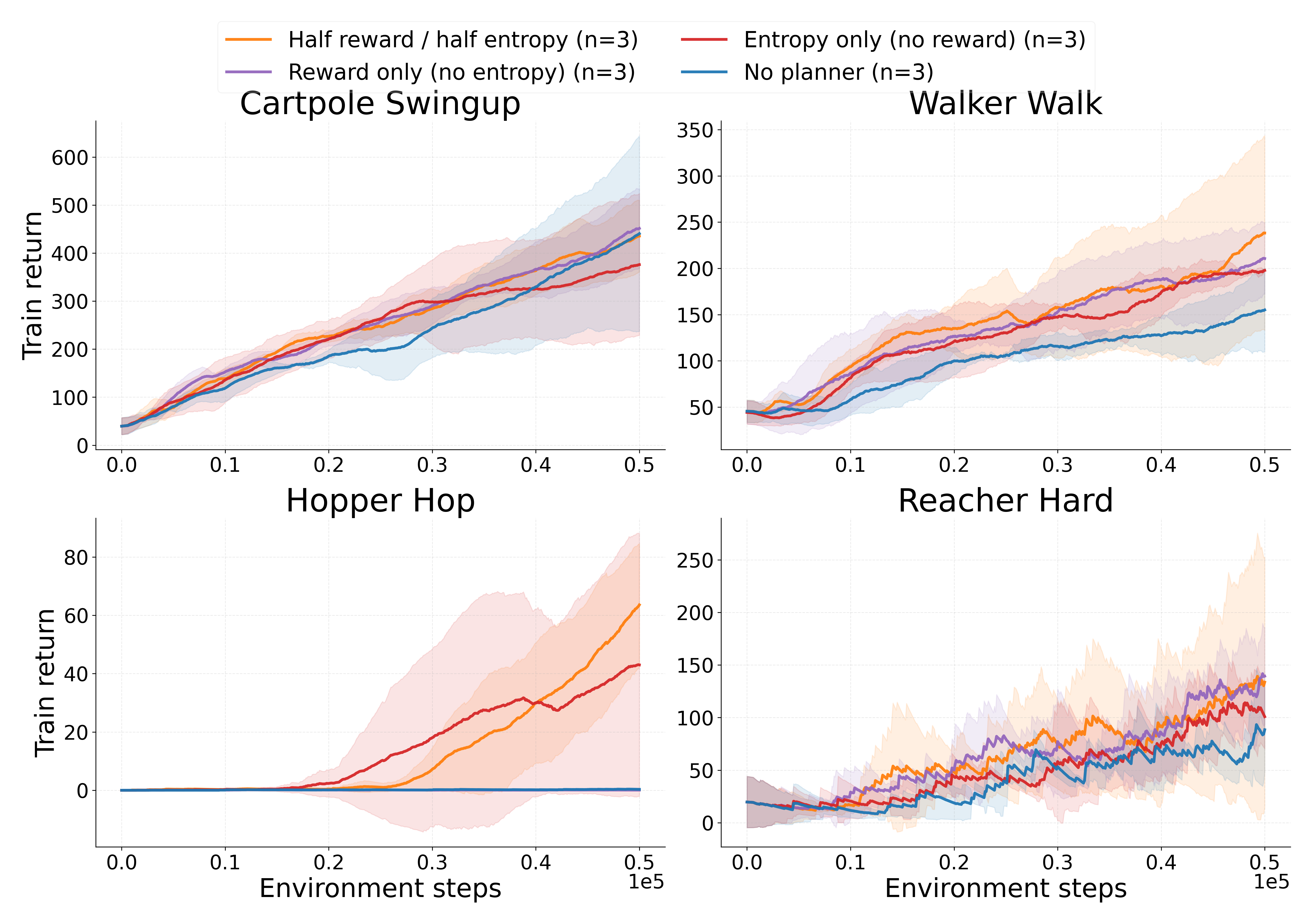}
        \caption{\rev{DMC entropy/reward ablation. Hopper notably fails without entropy. A larger version of this plot is provided in Figure~\ref{fig:dmc_entropy_ablate_app} for clarity.}}.
        \label{fig:dmc_entropy_ablate}
    \end{subfigure}
    \caption{Entropy/reward ablations for Crafter (left) and DMC (right). We compare training the meta-policy with entropy-only, reward-only, and a 50/50 mixture of both. In both domains, entropy-only and reward-only training of the meta-policy are comparable and outperform no-planner, suggesting robustness to the precise meta-reward weighting.}
    \label{fig:entropy_reward_ablate}
\end{figure}

\rev{Figures \ref{fig:train_length_ablate_objectnav} and \ref{fig:train_length_ablate_crafter} demonstrates the value of committing to plans over extended horizons. Myopic planning causes dithering in both the maze and Crafter environments, with the full planner outperforming MPC-style replanning by leverating plan commitment.} \rev{The mixed objective performs slightly worse than either pure objective in the Crafter env, as shown in Fig.~\ref{fig:crafter_entropy_ablate}, which we attribute to the two terms being only partially aligned in the environment: combining them can reduce selection contrast and dilute the meta-policy's advantage signal, which is why we use pure entropy as our training signal (for Crafter). Importantly, the comparable performance of entropy-only and reward-only suggests that, under actor-guided proposals, reward and predictive uncertainty often point to similar futures. Entropy therefore offers a stable default planning signal when rewards are not available, while remaining competitive when dense rewards are available. The lower variance we observe with entropy is consistent with uncertainty being a more directly model-aligned quantity than predicted reward. In Fig.~\ref{fig:dmc_entropy_ablate} we see all three planning objectives performing largely the same, except in hopper where entropy-less objectives fail. Hopper has one of the highest reward scarcities in our experimental testbench, without an effective exploration signal it can be hard to perform well in it. Additional sensitivity sweeps over candidate count and PPO horizon are shown in Appendix~\ref{sec:appendix_ablation_sensitivity}.}
\subsection{Timing Analysis}
\rev{Our method adds inference-time overhead from generating imagined rollouts and training the meta-policy. With default settings ($H{=}16$, $N{=}256$ candidates), a single planning call costs $\approx 0.05$s across all regimes. The cost scales linearly with rollout horizon, while varying the number of candidates has only a small effect due to batched evaluation on GPU. Meta-policy training adds a further $\sim$8-10ms per PPO update at update frequency 32. Because replanning is commit-aware, these costs are amortized over multiple environment steps. Full timing tables and scaling sweeps are reported in Appendix~\ref{sec:appendix_timing}. We show in Appendix~\ref{sec:appendix_planner_metrics} that the replan rate drops as the agent trains, increasing planner efficiency and possibly reflecting the world model's error rate decreasing with training.}
\section{Limitations \& Conclusion}
\label{sec:Conclusion}
We note that an inherent limitation of our method is that the actor must be trained purely with world model generated states rather than through experience replay; this method biases collection of experiences towards high model entropy, leading to a distributional shift between the actor's policy and the actual behaviour policy. \rev{More broadly, we see inference-time entropy planning as a lightweight anticipatory layer for world-model agents. It requires no auxiliary reward learning and fits onto existing objectives. Retrospective intrinsic rewards remain valuable for shaping long-term state distributions, and integrating both signals is a natural direction for future work.}

\paragraph{Reproducibility Statement.}
We document datasets, preprocessing, and step-by-step training and evaluation procedures in the Experiments section. Random seeds used for all runs are listed in the experiments section. We will release the complete codebase on GitHub upon acceptance; in the meantime, the paper and appendix provide all details needed to reimplement our results. Configuration settings are located in Appendix \ref{sec:appendix_config}.

\bibliography{iclr2026_conference}
\bibliographystyle{iclr2026_conference}

\newpage
\appendix

\section{LLM / AI Tooling Disclosure}
\label{sec:appendix_ai}
We used AI-assisted tools during this project as follows.

\paragraph{Tools.}
\begin{itemize}
    \item \textbf{Cursor (AI coding assistant):} Used to generate boilerplate code, suggest refactorings, produce docstrings and unit-test skeletons, and surface API idioms. All produced code was reviewed, modified as needed, and verified by the authors.
    \item \textbf{ChatGPT (writing assistant):} Used for copy-editing, grammar and style suggestions, tightening wording, expanding or condensing paragraphs on request, and clarifying phrasing. We did not use it for ideation, technical contributions, or to generate substantive claims.
\end{itemize}

The authors reviewed and verified all AI-assisted outputs for correctness and originality, and accept full responsibility for the code and text included in the paper. Any code suggested by tools was tested and adapted to our setting before inclusion.

\section{\rev{MDP Formalism}}
\label{sec:appendix_mdp_formalism}

We briefly summarise the Markov decision process (MDP) and partially observable MDP (POMDP) formalisms to make the assumptions behind Dreamer and our planner explicit.

\paragraph{Markov decision process (MDP).}
An MDP is a tuple $\mathcal{M} = \langle \mathcal{S}, \mathcal{A}, p, r, \gamma, \rho_0 \rangle$, where
$\mathcal{S}$ is the set of environment states, $\mathcal{A}$ the set of actions,
$p(s_{t+1}\mid s_t, a_t)$ the transition dynamics,
$r(s_t,a_t)$ the expected immediate reward,
$\gamma\in(0,1)$ the discount factor, and
$\rho_0$ the initial-state distribution \citep{kaelbling1998pomdp}.
The Markov property, or the memoryless assumption, states that $(s_{t+1}, r_t)$ depend only on $(s_t,a_t)$. A stationary policy $\pi(a\mid s)$ induces trajectories by
$s_0\sim \rho_0$, $a_t\sim \pi(\cdot\mid s_t)$, $s_{t+1}\sim p(\cdot\mid s_t,a_t)$.
Following these trajectories, the standard RL objective is to maximise expected discounted return using the Bellman equation:
\begin{equation}
J(\pi) \;=\; \mathbb{E}_{\pi,p}\!\left[\sum_{t=0}^{\infty}\gamma^t r(s_t,a_t)\right].
\end{equation}

In many environments the agent does not observe $s_t$ directly. A POMDP extends the MDP to
$\mathcal{P} = \langle \mathcal{S}, \mathcal{A}, p, r, \mathcal{X}, Z, \gamma, \rho_0 \rangle$,
where $\mathcal{X}$ is an observation space and
$Z(x_t\mid s_t)$ is the observation (emission) model \citep{kaelbling1998pomdp}.
The process evolves as
\begin{align}
s_0 &\sim \rho_0, \\
a_t &\sim \pi(\cdot \mid x_{1:t}, a_{1:t-1}), \\
s_{t+1} &\sim p(\cdot\mid s_t,a_t), \\
x_{t+1} &\sim Z(\cdot\mid s_{t+1}),
\end{align}
so the policy must condition on the \emph{history} because $s_t$ is hidden. A sufficient statistic
for the history is the belief state
$b_t(s)=p(s_t=s \mid x_{1:t},a_{1:t-1})$, yielding an equivalent fully observed \emph{belief-MDP}.
We use the POMDP view throughout, since Dreamer operates from pixels and must infer latent state.

Dreamer’s RSSM can be interpreted as learning a compact belief representation for a POMDP:
the deterministic recurrent state $h_t$ summarises history, the prior $p_\phi(\rz_t\!\mid h_t)$ predicts
latent futures, and the posterior $q_\phi(\rz_t\!\mid h_t,x_t)$ refines this belief after observing $x_t$,
trained via the KL loss in Eq.~\eqref{eq:kl_loss}. In continual RL, \citet{khetarpal2022towards}
emphasise that both partial observability and nonstationarity can be modelled by augmenting the hidden
state with task/phase variables, i.e., treating continual learning as a (possibly changing) POMDP.
This perspective motivates using model uncertainty over latent state as a principled exploration signal:
high-entropy priors correspond to broad beliefs about unobserved environment factors, which our planner
seeks out in imagined futures.

\section{Planning Algorithm}
\label{sec:appendix_B}
\begin{algorithm}[H]
\caption{Entropy Seeking Anticipatory Planning}
\begin{algorithmic}[1]
\State \textbf{Input:} Observation $o_t$
\State Meta-policy computes discrete planning probability $p_t \in \{0, 0.06, 0.25, 0.56, 1\}$ (via squaring sampled values from $\{0, 0.25, 0.5, 0.75, 1.0\}$)
\State Sample $u_t \sim \mathcal{U}(0,1)$
\If{$u_t < p_t$}
    \State Greedy actor samples $C$ (256) candidate actions $a_1, \dots, a_C$ from $o_t$
    \For{$i = 1$ to $C$}
        \State Roll out trajectory $\tau_i$ of length $H$ (maximum rollout length, 16 here) using world model and greedy actor
        \State Compute $E_H = \frac{1}{H} \sum_{t=1}^{H} E_t^{(i)}$
    \EndFor
    \State Select trajectory $\tau_{best} = \arg\max E_H$
    \State Set plan to $\tau_{best}$
\EndIf
\State Continue interacting with environment; repeat planning check at next step
\end{algorithmic}
\end{algorithm}

\section{MiniWorld Environment and Reward Scheme}
\label{sec:appendix_C}
We extend the MiniWorld Maze environment\citep{MinigridMiniWorld23} with several task relevant augmentations. The maze environment is a 3 dimensional procedurally generated maze of size 8x8 (using recursive backtracking) where the agent can take continuous actions along three dimensions - forward/back (step size attenuated if moving backwards to encourage progress), strafe left/right, turn left/right. Each observation consists of a forward-facing RGB image of size (64x64x3). Each episode ends when the time limit (4096) is reached, or the three goal boxes have been found. Each training run samples a new maze structure every episode to prevent memorization. No regions of the maze are sectioned off from the rest of the maze and all the goal states are reachable.

To promote structured exploration, we introduce a porosity parameter that controls wall density: with probability $p$, wall segments are randomly removed during generation. This provides a tunable complexity gradient for navigation tasks by creating variable maze connectivity.

An auxiliary binary 2D map of size (64x64x3) that records agent visitation over the course of an episode has been concatenated to the observation. This map records visited coordinates as 1s whereas unvisited coordinates are kept at 0. The position of the agent and the direction it is looking in is also visible on the map. This serves as episodic spatial memory that enables agents to reason about coverage and connect their actions to the current observation. This mirrors plausible real-world capabilities that can be enacted through GPS tracking or odometry.

The reward function consists of three components: \begin{description}
\item[Exploration Reward:] A positive reward is granted when the agent visits a previously unvisited cell in its binary exploration map. 
The reward magnitude is proportional to the number of newly visited cells within a square region around the agent, the size of which is controlled by the \textit{blur} parameter, given by $b$. 
While this reward introduces non-Markovian dynamics by incorporating visitation history, the inclusion of a binary map in the observation allows memory-less model-free agents such as PPO to perform effectively in this environment.
\[
\begin{aligned}
\text{exploration reward} =
\begin{cases}
\displaystyle \frac{\Delta_t}{b^2} & \text{if } b > 1 \\
\Delta_t & \text{otherwise}
\end{cases}
\\ 
\text{where } \Delta_t = \text{number of newly explored cells at time } t
\end{aligned}
\]
\item[Proximity Reward:] A smoothly decaying signal is emitted by each goal object, with exponentially scaled rewards given when the agent is within an $x$-unit radius. This mimics real-world analogs such as bluetooth signals or radio signals for search and rescue, animal noises for ecological monitoring, or semantic hints for more advanced exploration. This reward takes the form of two bars in the center of the image - if the agent is near a goal box of a particular colour (red, green, or blue), the bars will turn that colour with intensity varying with distance.
\[
\begin{aligned}
\text{Proximity reward} = 
\begin{cases}
0 & \text{if } \Delta < 0 \text{ or } \Delta > 10 \\
(10 - \Delta)^2 \cdot p_{mul} & \text{otherwise}
\end{cases}
\\
\text{where } \Delta = \text{dist} - (r_{\text{agent}} + r_{\text{box}} + s) \text{ and } p_{mul} = 0.03
\end{aligned}
\]
\item[Goal Reward:] The agent gets a reward for moving into a coloured box. It gets $50$ per box and then $150$ when it gets the third box.
\end{description}
Thus the overall reward is composed of these three elements summed onto the baseline of -10. The lower limit of reward gained in an episode is $-T$ where $T$ is the time limit, and the upper limit is 0.

\section{Maze Images}
\label{sec:appendix_F}
\begin{figure}[tb]
    \centering
    \begin{subfigure}{0.3\textwidth}
        \centering
        \includegraphics[width=\linewidth]{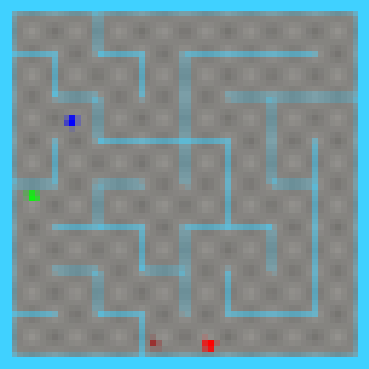}
        \caption{Porosity 0.0}
    \end{subfigure}
    \hfill
    \begin{subfigure}{0.3\textwidth}
        \centering
        \includegraphics[width=\linewidth]{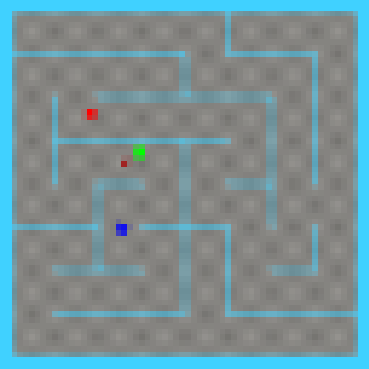}
        \caption{Porosity 0.1}
    \end{subfigure}
    \hfill
    \begin{subfigure}{0.3\textwidth}
        \centering
        \includegraphics[width=\linewidth]{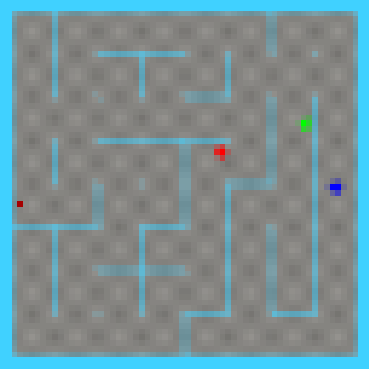}
        \caption{Porosity 0.2}
    \end{subfigure}


    \begin{subfigure}{0.3\textwidth}
        \centering
        \includegraphics[width=\linewidth]{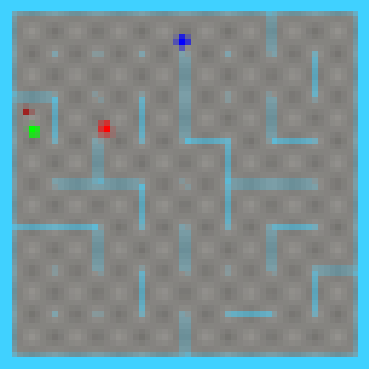}
        \caption{Porosity 0.4}
    \end{subfigure}
    \hfill
    \begin{subfigure}{0.3\textwidth}
        \centering
        \includegraphics[width=\linewidth]{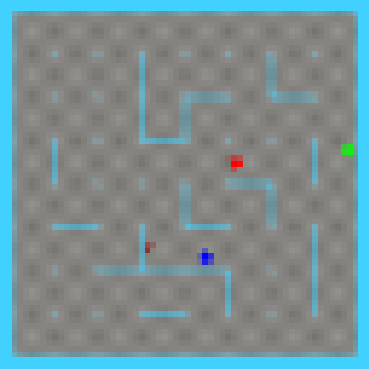}
        \caption{Porosity 0.6}
    \end{subfigure}
    \hfill
    \begin{subfigure}{0.3\textwidth}
        \centering
        \includegraphics[width=\linewidth]{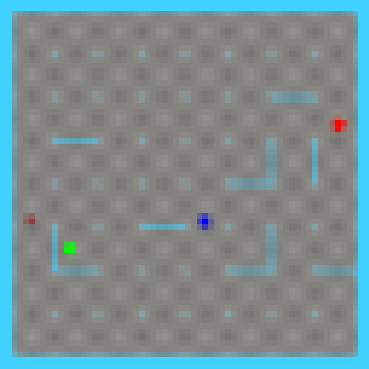}
        \caption{Porosity 0.8}
    \end{subfigure}

    \caption{Top-down maze layouts at selected porosity levels. Higher porosity values remove more internal walls, increasing openness and reducing planning difficulty.}
    \label{fig:maze_porosity_grid}
\end{figure}

To visualize the effect of varying porosity on maze complexity, we provide top-down views of generated mazes at increasing porosity levels, see Fig.~\ref{fig:maze_porosity_grid}. As porosity increases, more internal walls are removed, resulting in more open environments. These top-down maps reflect the structural differences that influence planning difficulty.

\begin{figure}[htb] 
    \centering 
        \begin{subfigure}{0.48\textwidth} \centering \includegraphics[width=\linewidth]{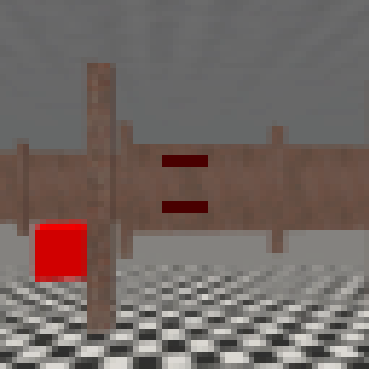} \caption{Agent’s visual observation} \label{fig:example_obs} \end{subfigure} 
        \hfill 
        \begin{subfigure}{0.48\textwidth} \centering \includegraphics[width=\linewidth]{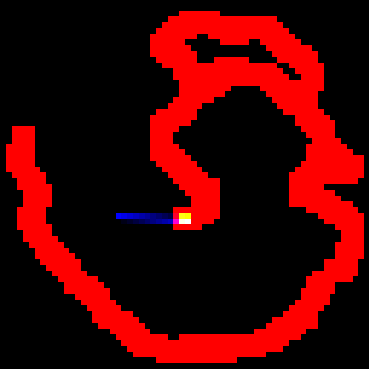} 
        \caption{Agent’s map observation} \label{fig:example_map} \end{subfigure} \caption{Image of what the agent perceives - the visual observation (left) and of the map observation (right).} \label{fig:example_view} 
\end{figure}

To contextualize the agent’s perspective within these mazes, we also provide an example of the full map layout and a corresponding visual observation seen by the agent, as given in Figure \ref{fig:example_view}.

\section{\rev{Timing and Compute Overhead}}
\label{sec:appendix_timing}

This appendix reports detailed timing for our planning module and the PPO-based meta-policy.
All timings were measured on the same hardware used for the main experiments, with the default
planner unless otherwise stated ($H{=}16$, $N{=}256$ candidates).

\subsection{Planning Call Latency}

Table~\ref{tab:timing_default} reports the mean wall-clock cost of a single imagination-and-score
planning call under default settings. Across Maze, Crafter, and DMC-Vision, a planning call is
consistently $\approx$50ms.

\begin{table}[t]
\centering
\small
\begin{tabular}{lccc}
\toprule
Environment & $H{=}16$, $N{=}256$ mean (s) & std (s) & notes \\
\midrule
MiniWorld Maze & 0.0463 & 0.0008 & continuous actions \\
Crafter        & 0.0493 & 0.0010 & discrete actions \\
DMC-Vision     & 0.0478 & 0.0007 & continuous actions \\
\bottomrule
\end{tabular}
\caption{Default planning-call latency. Values are mean $\pm$ std seconds per call.}
\label{tab:timing_default}
\end{table}

\subsection{Scaling with Horizon and Candidate Count}

Planning cost scales approximately linearly with the rollout horizon $H$:
halving $H$ from 16 to 8 roughly halves latency in all environments (e.g.,
Maze: 0.046s $\rightarrow$ 0.022s; Crafter: 0.049s $\rightarrow$ 0.025s; DMC: 0.048s $\rightarrow$ 0.024s).
By contrast, varying the number of candidates $N$ has only a small effect on wall-clock time,
because candidates are evaluated in a single batched GPU forward pass.
As a concrete illustration, replanning every other step adds $\approx 0.05\text{s}\times 50{,}000 \approx 2{,}500\text{s}$ ($\sim$42 minutes) per 100k environment steps.

For completeness, Tables~\ref{tab:timing_maze_matrix}--\ref{tab:timing_dmc_matrix}
show the full horizon--choices timing matrices.

\begin{table}[t]
\centering
\scriptsize
\begin{tabular}{lcccccccc}
\toprule
$H$ / $N$ & 256 & 128 & 64 & 32 & 16 & 8 & 4 & 2 \\
\midrule
16 & 0.0463 & 0.0464 & 0.0457 & 0.0445 & 0.0434 & 0.0423 & 0.0419 & 0.0446 \\
8  & 0.0225 & 0.0230 & 0.0227 & 0.0223 & 0.0216 & 0.0212 & 0.0209 & 0.0212 \\
4  & 0.0114 & 0.0116 & 0.0115 & 0.0114 & 0.0110 & 0.0106 & 0.0107 & 0.0109 \\
2  & 0.0058 & 0.0059 & 0.0059 & 0.0058 & 0.0056 & 0.0054 & 0.0054 & 0.0055 \\
\bottomrule
\end{tabular}
\caption{MiniWorld Maze planning-call timings (seconds per call). Means shown; stds are $<10^{-3}$s except for $N{=}2$ due to measurement noise.}
\label{tab:timing_maze_matrix}
\end{table}

\begin{table}[t]
\centering
\scriptsize
\begin{tabular}{lcccccccc}
\toprule
$H$ / $N$ & 256 & 128 & 64 & 32 & 16 & 8 & 4 & 2 \\
\midrule
16 & 0.0493 & 0.0495 & 0.0488 & 0.0479 & 0.0462 & 0.0450 & 0.0445 & 0.0471 \\
8  & 0.0246 & 0.0249 & 0.0244 & 0.0241 & 0.0232 & 0.0228 & 0.0225 & 0.0227 \\
4  & 0.0123 & 0.0125 & 0.0123 & 0.0122 & 0.0117 & 0.0115 & 0.0114 & 0.0117 \\
2  & 0.0063 & 0.0064 & 0.0063 & 0.0062 & 0.0060 & 0.0060 & 0.0059 & 0.0059 \\
\bottomrule
\end{tabular}
\caption{Crafter planning-call timings (seconds per call).}
\label{tab:timing_crafter_matrix}
\end{table}

\begin{table}[t]
\centering
\scriptsize
\begin{tabular}{lcccccccc}
\toprule
$H$ / $N$ & 256 & 128 & 64 & 32 & 16 & 8 & 4 & 2 \\
\midrule
16 & 0.0478 & 0.0485 & 0.0479 & 0.0467 & 0.0457 & 0.0443 & 0.0461 & 0.0445 \\
8  & 0.0240 & 0.0243 & 0.0239 & 0.0235 & 0.0228 & 0.0225 & 0.0221 & 0.0224 \\
4  & 0.0121 & 0.0122 & 0.0121 & 0.0119 & 0.0115 & 0.0114 & 0.0111 & 0.0113 \\
2  & 0.0062 & 0.0063 & 0.0062 & 0.0061 & 0.0059 & 0.0058 & 0.0057 & 0.0059 \\
\bottomrule
\end{tabular}
\caption{DMC-Vision planning-call timings (seconds per call).}
\label{tab:timing_dmc_matrix}
\end{table}

\subsection{Meta-policy (PPO) Sequence Processing}

The meta-policy is trained using PPO on sequences of length $L$.
Table~\ref{tab:ppo_seq_timing} reports the wall-clock cost of processing a batch of sequences
for different $L$. At the default $L{=}8$, PPO sequence processing costs $\approx$8--10ms per update,
which is small compared to the planning-call cost.

\begin{table}[t]
\centering
\small
\begin{tabular}{lcccc}
\toprule
Environment & $L{=}16$ & $L{=}8$ & $L{=}4$ & $L{=}2$ \\
\midrule
Maze    & 0.0261 & 0.0099 & 0.0092 & 0.0043 \\
Crafter & 0.0234 & 0.0083 & 0.0044 & 0.0041 \\
DMC     & 0.0240 & 0.0095 & 0.0048 & 0.0077 \\
\bottomrule
\end{tabular}
\caption{Meta-policy PPO sequence processing time (seconds per update) as a function of sequence length $L$.}
\label{tab:ppo_seq_timing}
\end{table}

\subsection{Environment Throughput and CPU Bottlenecks (MiniWorld)}

We observe that MiniWorld throughput is more CPU-bound than GPU-bound under our setup.
On otherwise comparable systems, 350k training steps took 16 hours on an RTX 5090 with an
Intel i9-14900K (24 cores, 6.0\,GHz max), compared to 24 hours on an RTX 4090 with an
i7-13700K (16 cores, 5.4\,GHz max), and 36 hours on an RTX 4070 Ti with a Ryzen 9 5900X
(12 cores, 4.8\,GHz max). This suggests that CPU core count and clock speed significantly
influence environment-step throughput in these RL environments (we found similar trends with the other two environments).

\section{\rev{Ablation and Sensitivity Analysis}}
\label{sec:appendix_ablation_sensitivity}

This appendix provides additional ablations and sensitivity sweeps referenced in the main text. We focus on (i) the meta-reward used to train the meta-policy, and (ii) robustness to planner hyperparameters such as the number of candidate rollouts and the commitment (sequence) horizon. Unless otherwise stated, each curve reports the mean with shaded $\pm 95\%$ confidence intervals computed as $2\times$SEM across seeds.

\subsection{Meta-reward (entropy vs.\ reward) ablations}
\label{sec:appendix_entropy_ablate}

\begin{figure*}[t]
    \centering
    \includegraphics[width=0.95\linewidth]{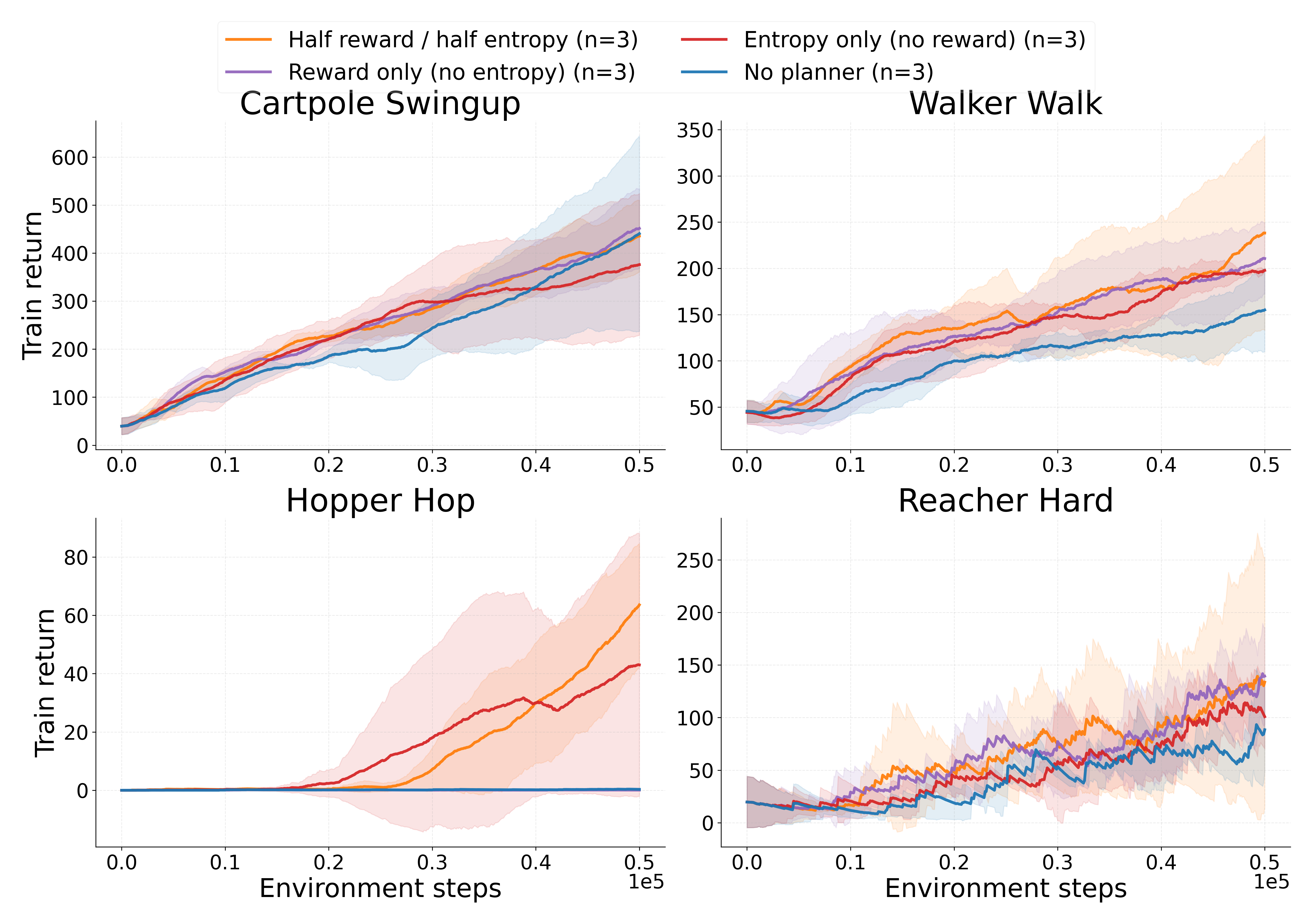}
    \caption{DMC-Vision meta-reward ablation. We vary the meta objective:
    entropy-only, reward-only, and a 50/50 mixture. Entropy-only and reward-only are comparable
    (reward-only slightly noisier), while the mixed objective is not consistently better,
    supporting robustness to $r_{\text{meta}}$ weighting.}
    \label{fig:dmc_entropy_ablate_app}
\end{figure*}

The main paper reports Crafter meta-reward ablations in Fig.~\ref{fig:crafter_entropy_ablate}.
Here we provide the corresponding ablations on representative DMC-Vision tasks
(Fig.~\ref{fig:dmc_entropy_ablate_app}). We vary the components of $r_{\text{meta}}$ between
entropy-only, reward-only, and a 50/50 mixture. As shown in Fig.~\ref{fig:dmc_entropy_ablate_app},
entropy-only and reward-only yield broadly similar learning trends across tasks, with reward-only
often exhibiting slightly higher variance. This is expected given the small number of seeds and the
fact that Dreamer’s reward predictor can be noisier than the world-model uncertainty signal. 
The mixed objective is not consistently better and in several cases is slightly worse, suggesting
that entropy and reward are only partially aligned in these tasks; mixing them can dilute the
planner's anticipatory drive without adding a clearer control signal. Overall, the ablation supports
our claim that the method is robust to the entropy reward weighting and does not require careful tuning.

\subsection{Sensitivity to candidate count}
\label{sec:appendix_sensitivity_candidates}

\begin{figure*}[t]
    \centering
    \begin{subfigure}[t]{0.49\linewidth}
        \centering
        \includegraphics[width=\linewidth]{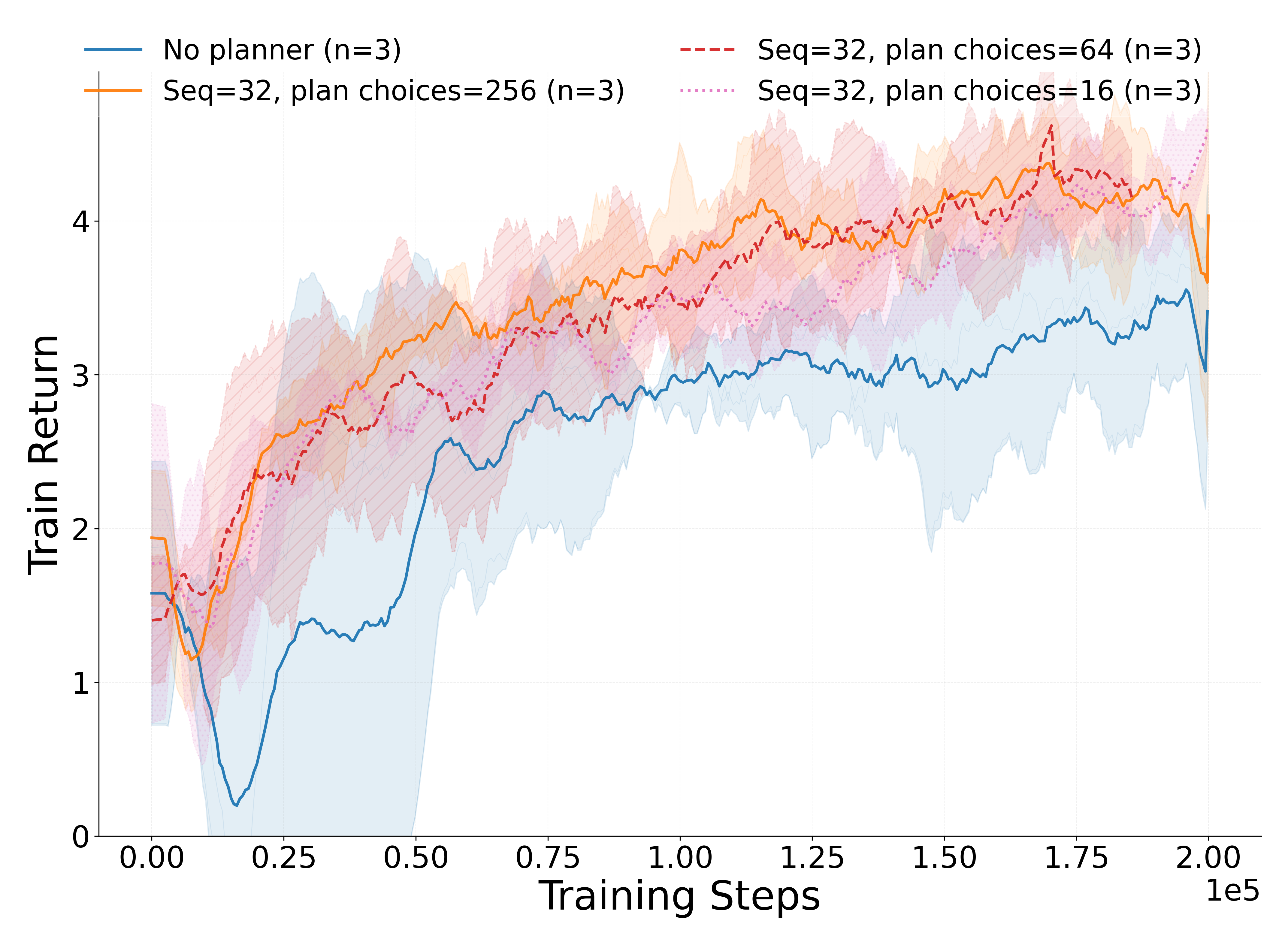}
        \caption{Candidate count sweep $N$ (plan choices).}
        \label{fig:crafter_sens_planchoices}
    \end{subfigure}\hfill
    \begin{subfigure}[t]{0.49\linewidth}
        \centering
        \includegraphics[width=\linewidth]{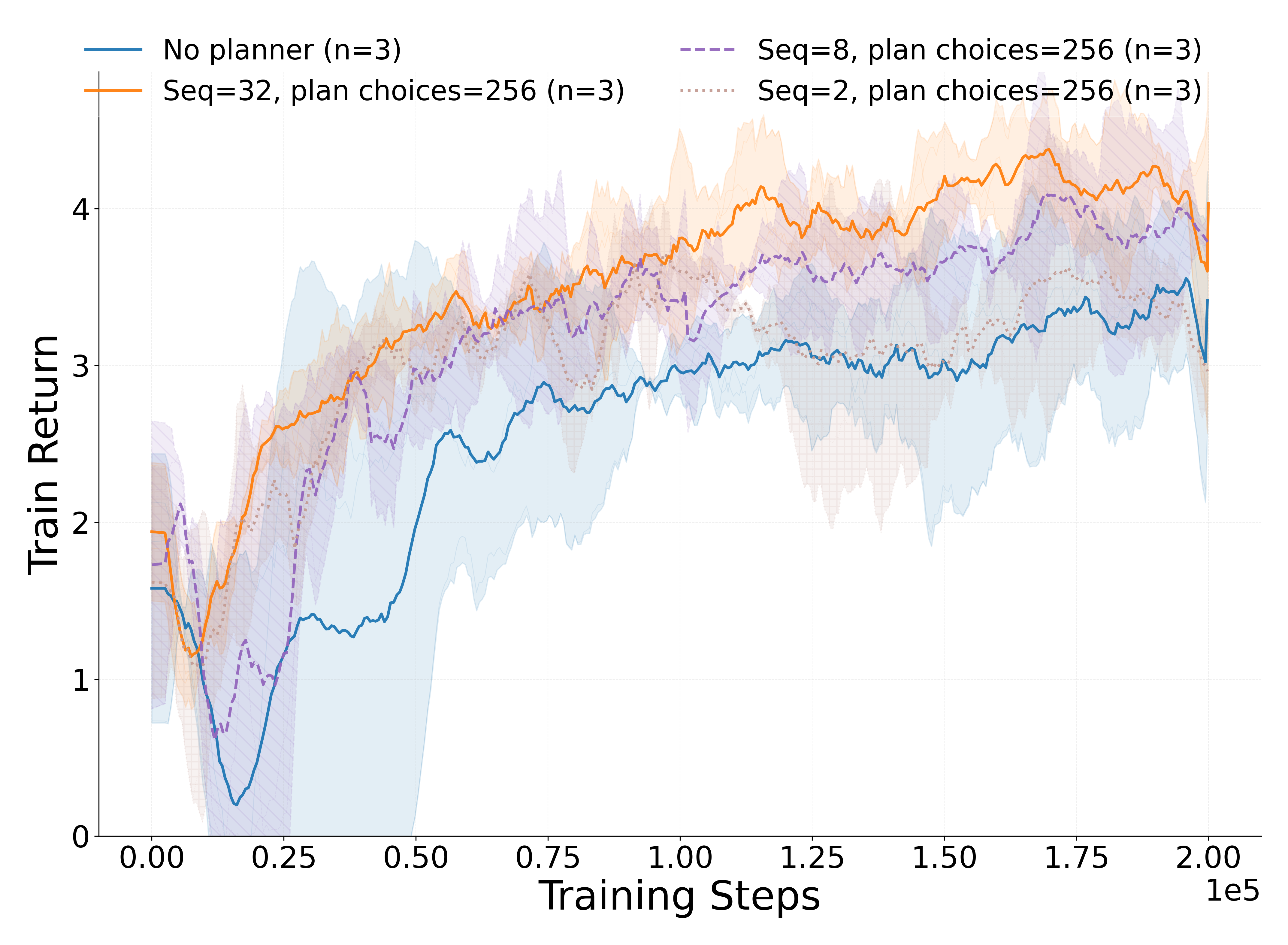}
        \caption{Commitment horizon sweep (seq length).}
        \label{fig:crafter_sens_seq}
    \end{subfigure}
    \caption{Crafter sensitivity sweeps. Left: varying the number of imagined candidates $N$ per replanning step.
    Right: varying the commitment / replanning horizon (seq). Curves show mean return with shaded
    $\pm 95\%$ confidence intervals computed as $2\times$SEM across 5 seeds.}
    \label{fig:crafter_sensitivity}
\end{figure*}

\begin{figure*}[t]
    \centering
    \includegraphics[width=0.98\linewidth]{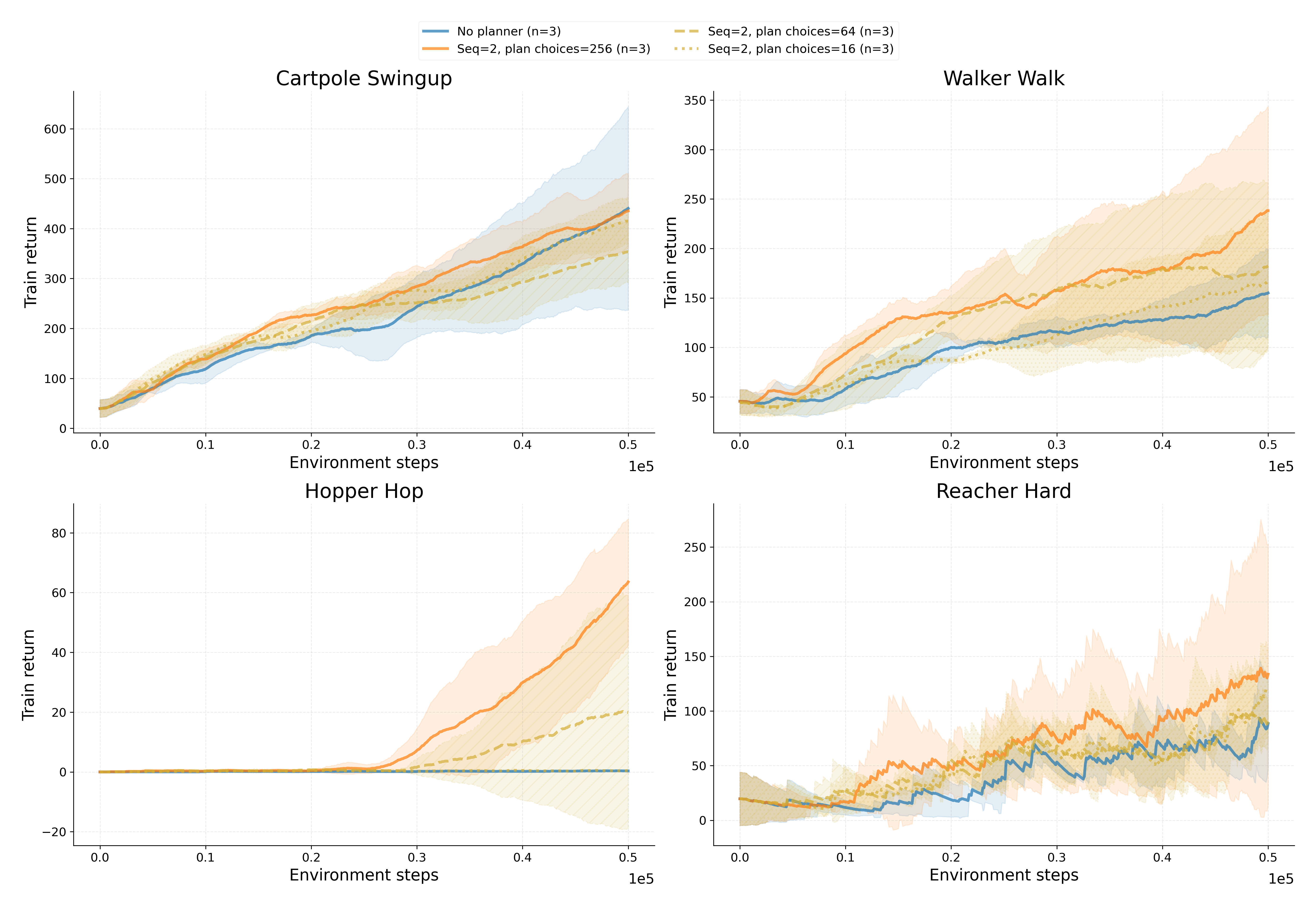}
    \caption{DMC-Vision sensitivity to candidate count $N$ (plan choices) across four tasks
    (\texttt{cartpole\_swingup}, \texttt{walker\_walk}, \texttt{hopper\_hop}, \texttt{reacher\_hard}).
    Each panel reports mean return with shaded $\pm 95\%$ confidence intervals ($2\times$SEM).}
    \label{fig:dmc_sens_planchoices}
\end{figure*}

We sweep the number of actor-guided candidates $N$ evaluated per replanning step.
Crafter sensitivity to candidate count is shown in the left panel of
Fig.~\ref{fig:crafter_sensitivity} (see Fig.~\ref{fig:crafter_sens_planchoices}),
while DMC-Vision sensitivity is shown in Fig.~\ref{fig:dmc_sens_planchoices}.
In both regimes, gains persist across a wide span of candidate counts, and the default
($N{=}256$) lies in a stable region. Very small candidate sets (e.g., $N{=}16$) can reduce
performance, consistent with the planner having fewer high-return futures to choose among.
Because candidates are generated by a greedy actor, trajectories are partially redundant,
so moderate oversampling is expected. Importantly, these sweeps align with the timing analysis
(Appendix~\ref{sec:appendix_timing}): increasing $N$ has only a modest impact on wall-clock cost
because candidate evaluation is batched on the GPU.

\subsection{Sensitivity to commitment / meta horizon}
\label{sec:appendix_sensitivity_seq}

\begin{figure*}[t]
    \centering
    \includegraphics[width=0.98\linewidth]{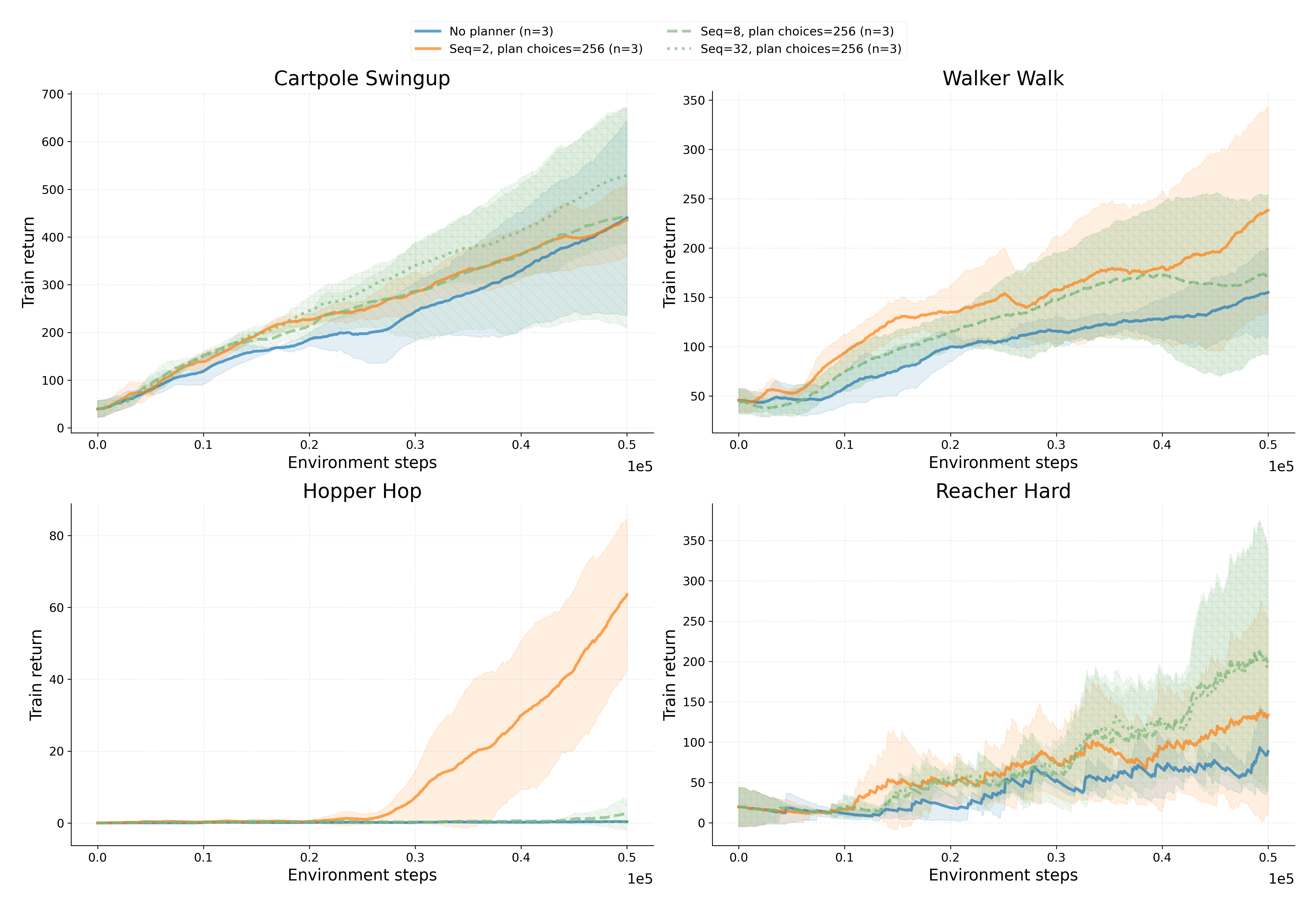}
    \caption{DMC-Vision sensitivity to commitment horizon (seq length) across four tasks
    (\texttt{cartpole\_swingup}, \texttt{walker\_walk}, \texttt{hopper\_hop}, \texttt{reacher\_hard}).
    Each panel reports mean return with shaded $\pm 95\%$ confidence intervals ($2\times$SEM).}
    \label{fig:dmc_sens_seq}
\end{figure*}

We sweep the commitment horizon (meta PPO sequence length / replanning interval, denoted ``seq'').
Crafter results are shown in the right panel of Fig.~\ref{fig:crafter_sensitivity}
(see Fig.~\ref{fig:crafter_sens_seq}), and DMC-Vision results are shown in
Fig.~\ref{fig:dmc_sens_seq}. Short-to-moderate commitment typically yields the most reliable gains,
matching the main finding that commitment reduces dithering while retaining flexibility to replan.
Longer commitment can occasionally help (e.g., for smoother dynamics), but is less stable overall.
These trends also explain the practical overhead reduction discussed in the timing analysis
(Appendix~\ref{sec:appendix_timing}): longer average commitment amortizes replanning cost
over more environment steps.

\section{\rev{Planner Metrics}}
\label{sec:appendix_planner_metrics}

This section reports diagnostics that characterise the learned meta-policy and help interpret the
practical behaviour of commit-aware replanning. Both plots aggregate quantities per episode and then average across seeds (and environment collections): the solid curve is the mean, the shaded region denotes a single standard deviation, averaged in the same way as the mean but calculated per episode, and the dotted curve shows the average episode-wise maximum, highlighting occasional extreme behaviours, or flexibility. These metrics also connect directly to the compute overhead discussion in Appendix~\ref{sec:appendix_timing}.

Figure~\ref{fig:plan_prob} plots the probability that the meta-policy chooses to replan at a given
environment step. Across domains, the meta-policy quickly moves away from “always replan” and
stabilises at a selective regime: Crafter gradually increases from roughly $0.5$ to around
$0.6$–$0.7$ over training, while Maze and DMC stabilise between $\sim0.6$ and $\sim0.7$. The dotted
average-maximum trace remains close to~$1.0$, showing that some episodes briefly approach near-always
replanning, but the mean behaviour does not collapse to this regime. Overall, the learned policy
maintains the intended commit-aware behaviour while retaining flexibility to replan more aggressively
when needed, while proposing different planning regimes for different environments and tasks.

\begin{figure}[tb]
\centering
\includegraphics[width=0.95\linewidth]{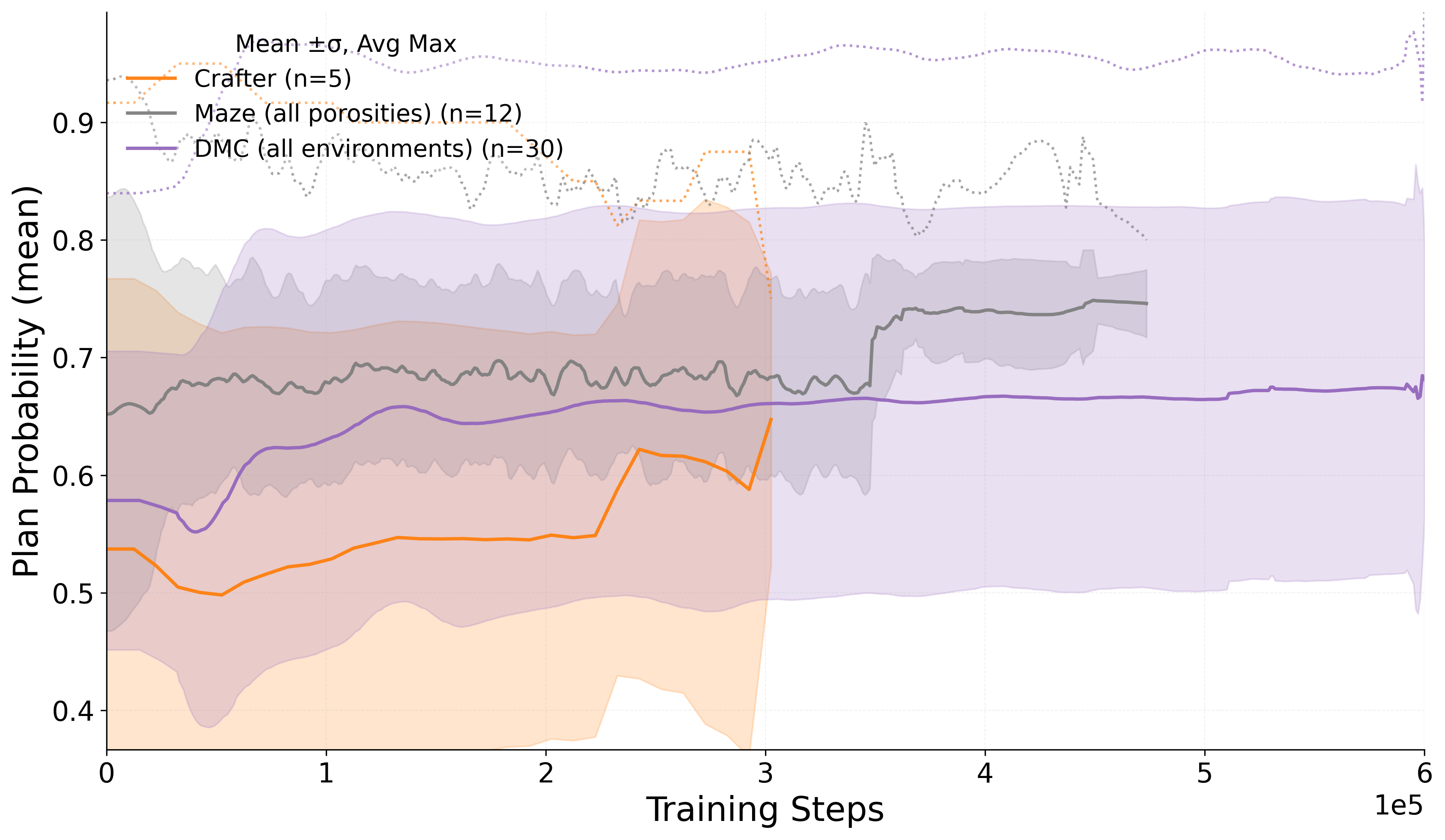}
\caption{Planning probability over training. Solid: mean across seeds; shaded band: 1 standard deviation averaged; dotted: average episode-wise maximum. The meta-policy settles into a selective
replanning regime, with replan probabilities typically between $\sim0.5$ and $\sim0.7$ rather than
replan-every-step behaviour.}
\label{fig:plan_prob}
\end{figure}

\begin{figure}[tb]
\centering
\includegraphics[width=0.95\linewidth]{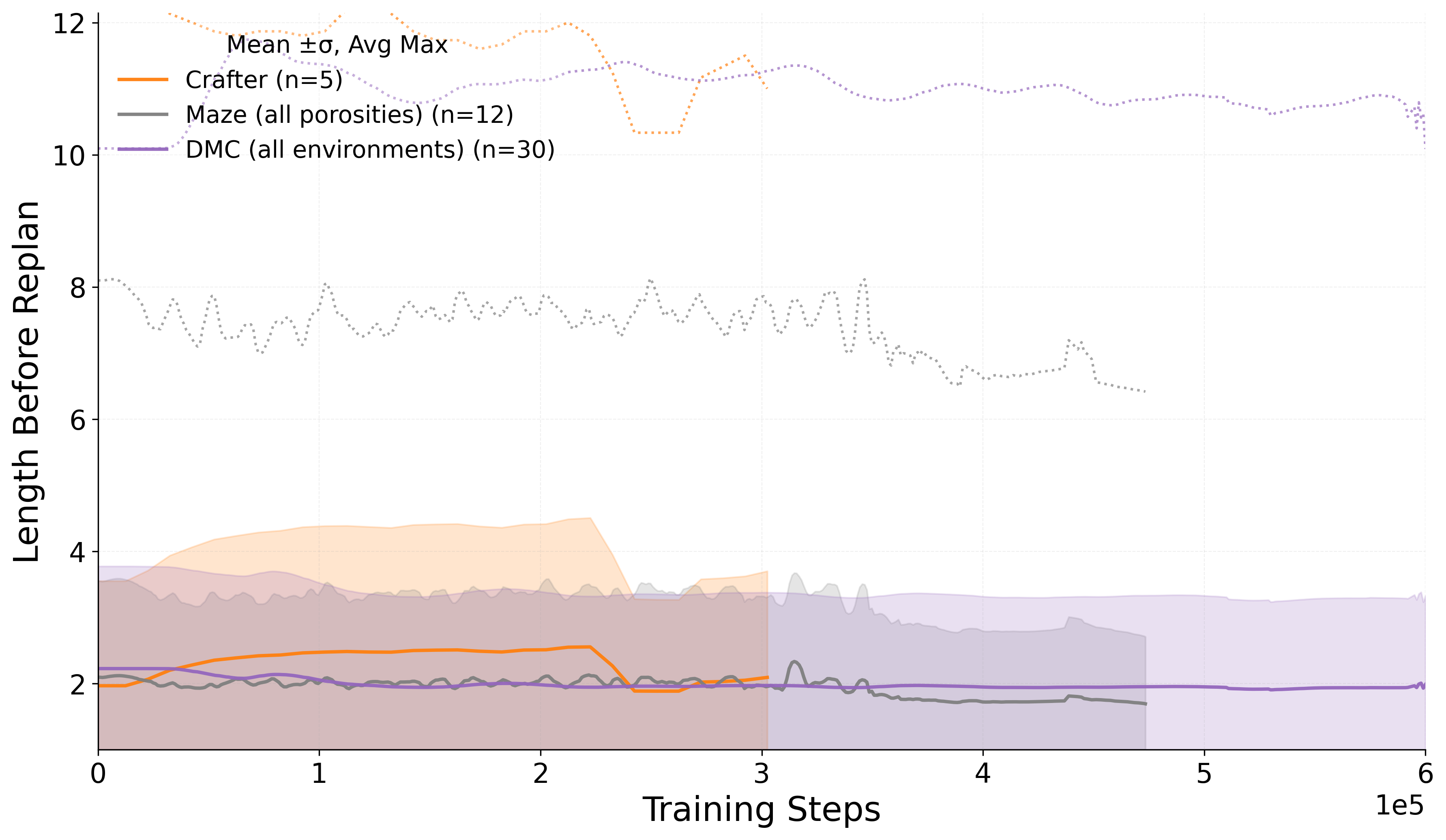}
\caption{Length before replanning. Solid: mean across seeds; shaded band: 1 standard deviation averaged; dotted: average episode-wise maximum. The planner usually commits for
$\sim2$--$3$ steps, with occasional long commitments.}
\label{fig:len_before_replan}
\end{figure}

Figure~\ref{fig:len_before_replan} reports the number of environment steps executed before the next
replanning event. Commitment lengths remain short on average: Maze and DMC typically commit for about
$\sim2$ steps before reconsidering, while Crafter shows slightly longer commitments early in training
($\sim2$–$3$ steps) before settling to a similar range. The dotted average-maximum curve, however,
remains much higher (around $\sim8$–$12$ steps depending on domain), indicating that some episodes
sustain substantially longer commitments even though typical behaviour favours frequent opportunities
to replan.

Appendix~\ref{sec:appendix_timing} reports an average cost of $\approx 0.05$s per planning call with
default settings. Figures~\ref{fig:plan_prob} and~\ref{fig:len_before_replan} show that replanning
occurs on only about half to two-thirds of environment steps, and that each plan is typically executed
for multiple steps before replanning. Thus the planning cost is amortised over committed rollouts,
reducing the effective overhead per step by roughly a factor of two relative to a worst-case
“replan every step” scenario. In expectation this corresponds to an added $\sim25$–$30$ms per step
(about 40–50 minutes per 100k steps), consistent with the wall-clock parity discussion in the timing
analysis. Commit-aware replanning therefore provides both behavioural benefits (reduced dithering) and
practical throughput gains by avoiding unnecessary imagined rollouts.
\section{\rev{Crafter achievement breakdown}}
\label{app:crafter_achievements}

\begin{figure*}[t]
  \centering
  \captionsetup{font=small}
  \begin{subfigure}[t]{0.32\linewidth}
    \centering\includegraphics[width=\linewidth]{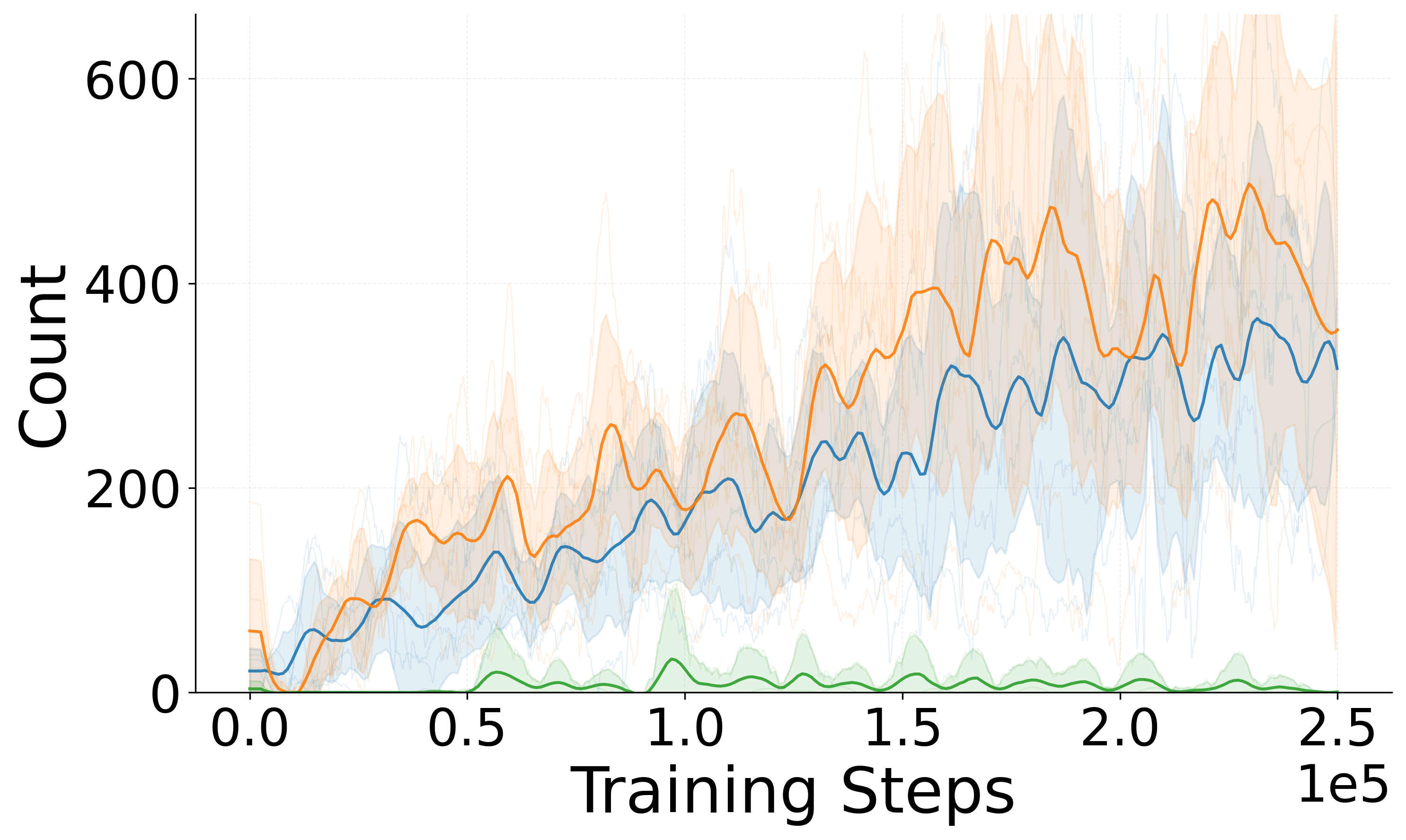}
    \caption{collect\_wood}
    \label{fig:crafter_collect_wood}
  \end{subfigure}\hfill
  \begin{subfigure}[t]{0.32\linewidth}
    \centering\includegraphics[width=\linewidth]{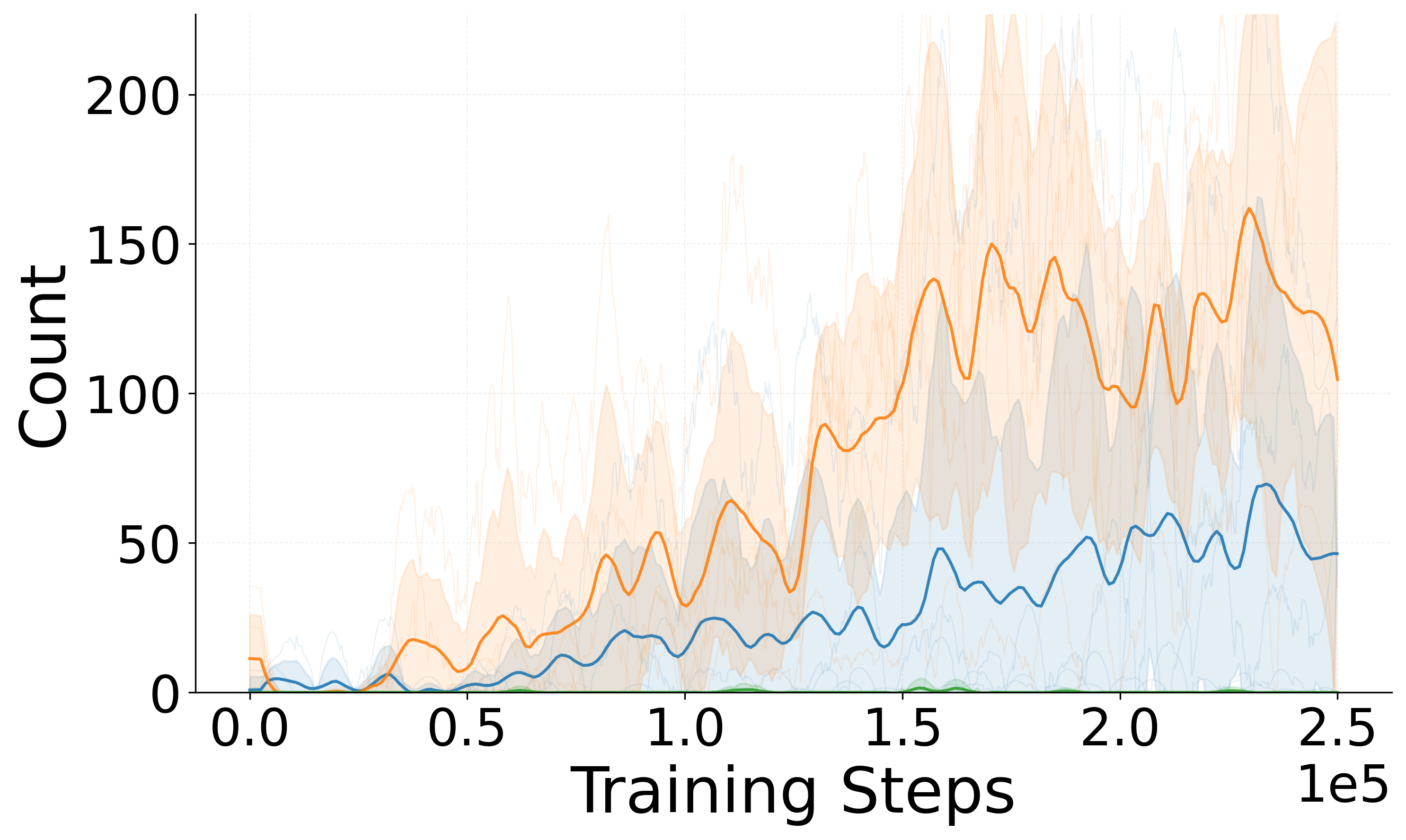}
    \caption{place\_table}
    \label{fig:crafter_place_table}
  \end{subfigure}\hfill
  \begin{subfigure}[t]{0.32\linewidth}
    \centering\includegraphics[width=\linewidth]{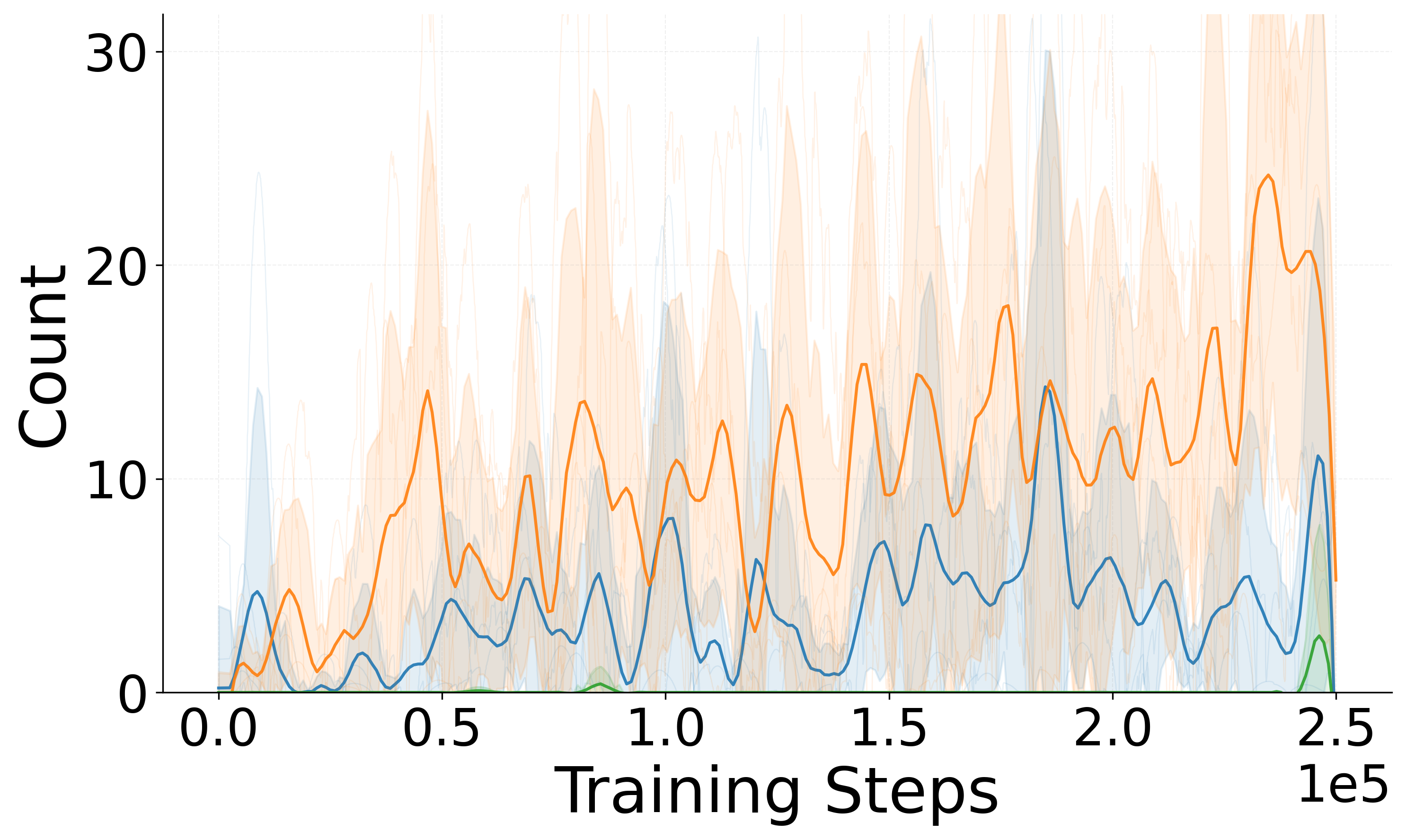}
    \caption{defeat\_zombie}
    \label{fig:crafter_defeat_zombie}
  \end{subfigure}

  \begin{subfigure}[t]{0.32\linewidth}
    \centering\includegraphics[width=\linewidth]{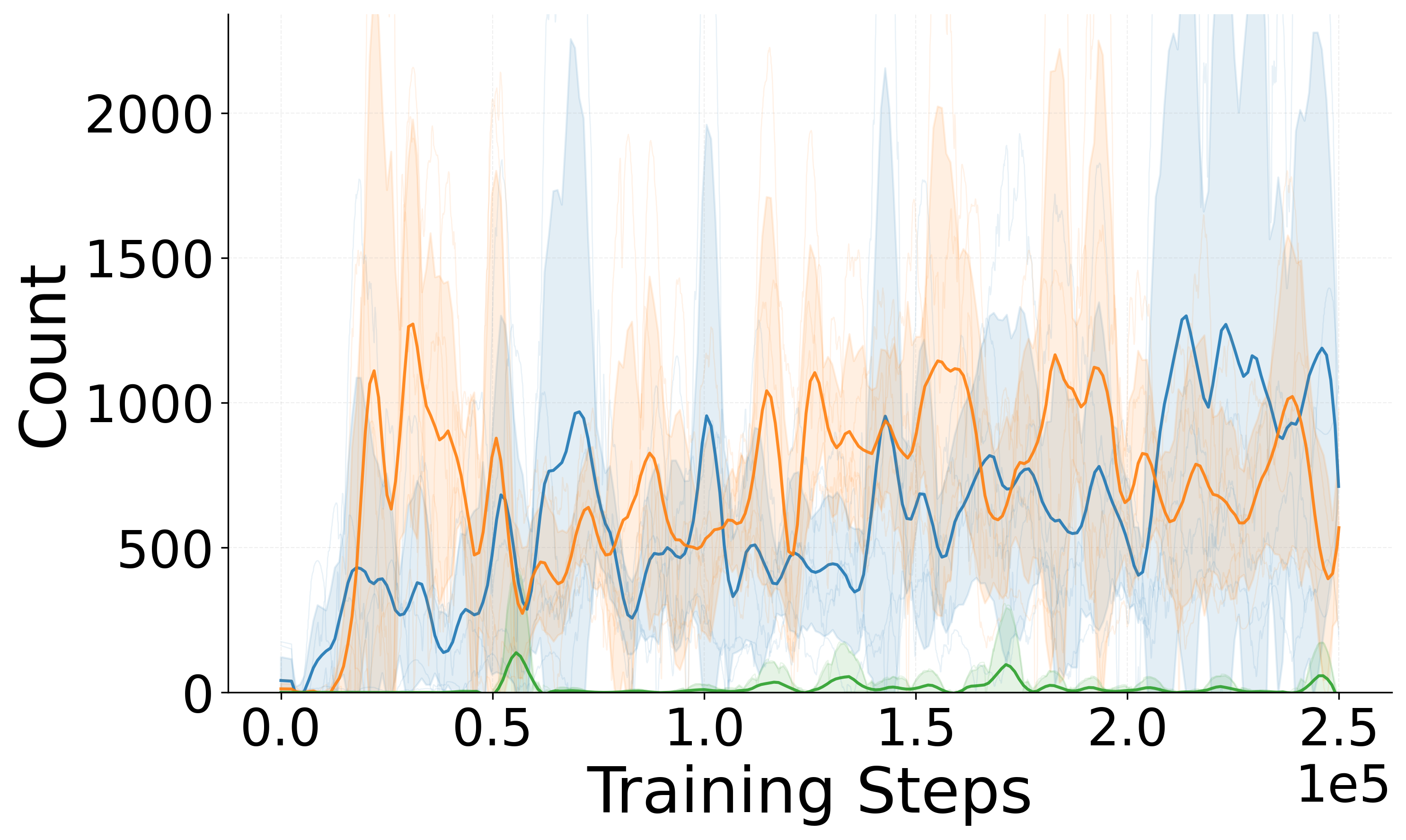}
    \caption{collect\_drink}
    \label{fig:crafter_collect_drink}
  \end{subfigure}\hfill
  \begin{subfigure}[t]{0.32\linewidth}
    \centering\includegraphics[width=\linewidth]{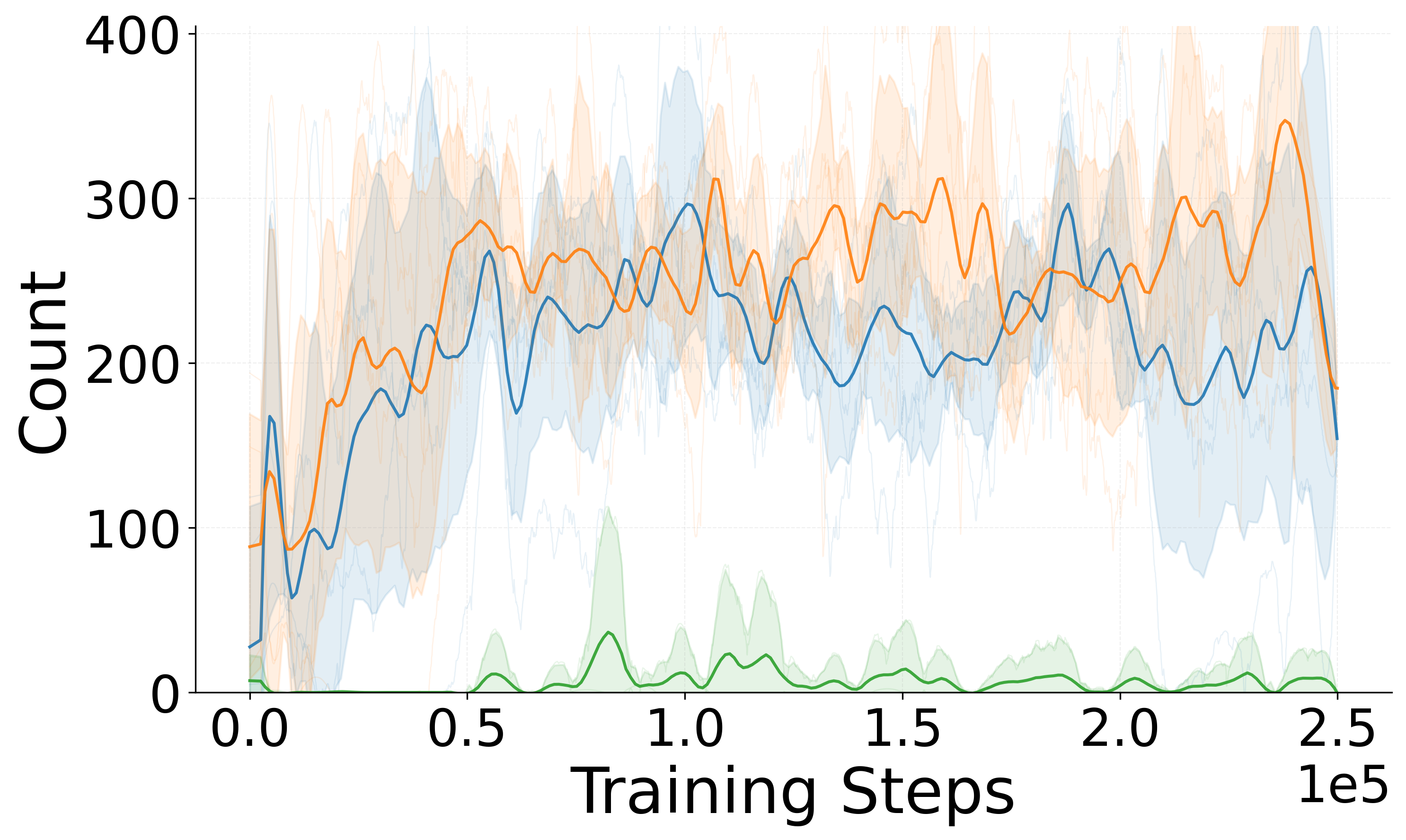}
    \caption{collect\_sapling}
    \label{fig:crafter_collect_sapling}
  \end{subfigure}\hfill
  \begin{subfigure}[t]{0.32\linewidth}
    \centering\includegraphics[width=\linewidth]{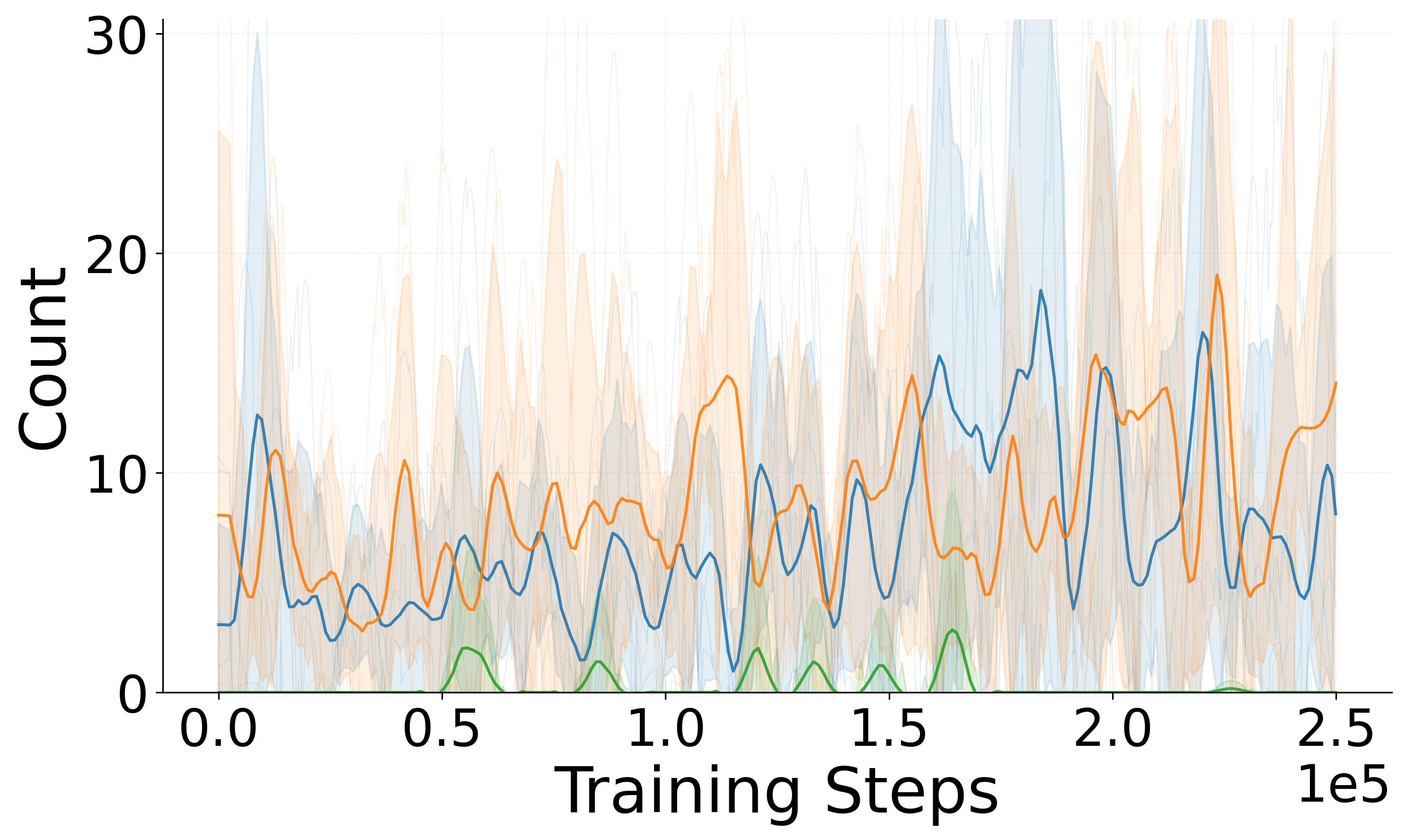}
    \caption{eat\_cow}
    \label{fig:crafter_eat_cow}
  \end{subfigure}

  \begin{subfigure}[t]{0.32\linewidth}
    \centering\includegraphics[width=\linewidth]{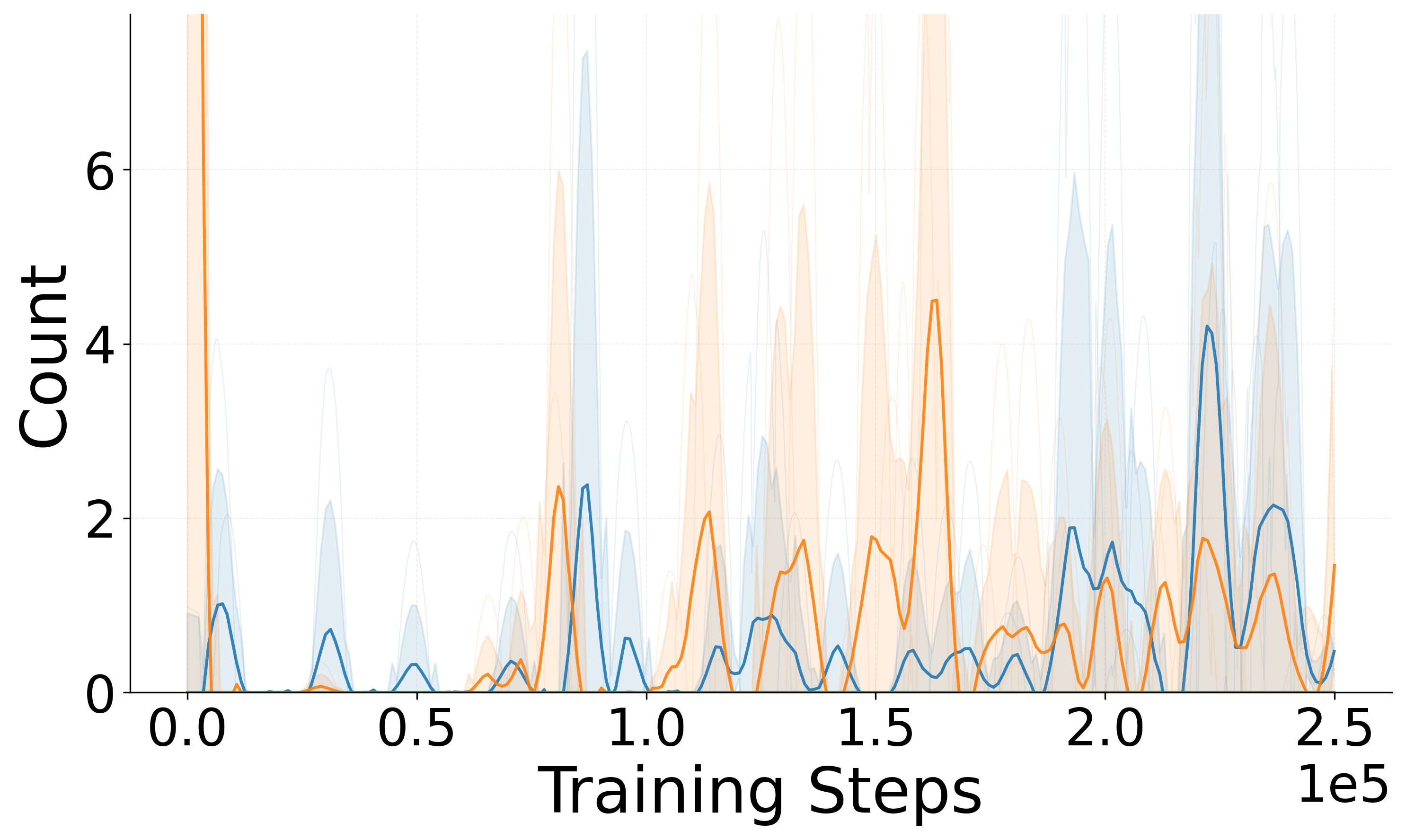}
    \caption{make\_wood\_pickaxe}
    \label{fig:crafter_make_wood_pickaxe}
  \end{subfigure}\hfill
  \begin{subfigure}[t]{0.32\linewidth}
    \centering\includegraphics[width=\linewidth]{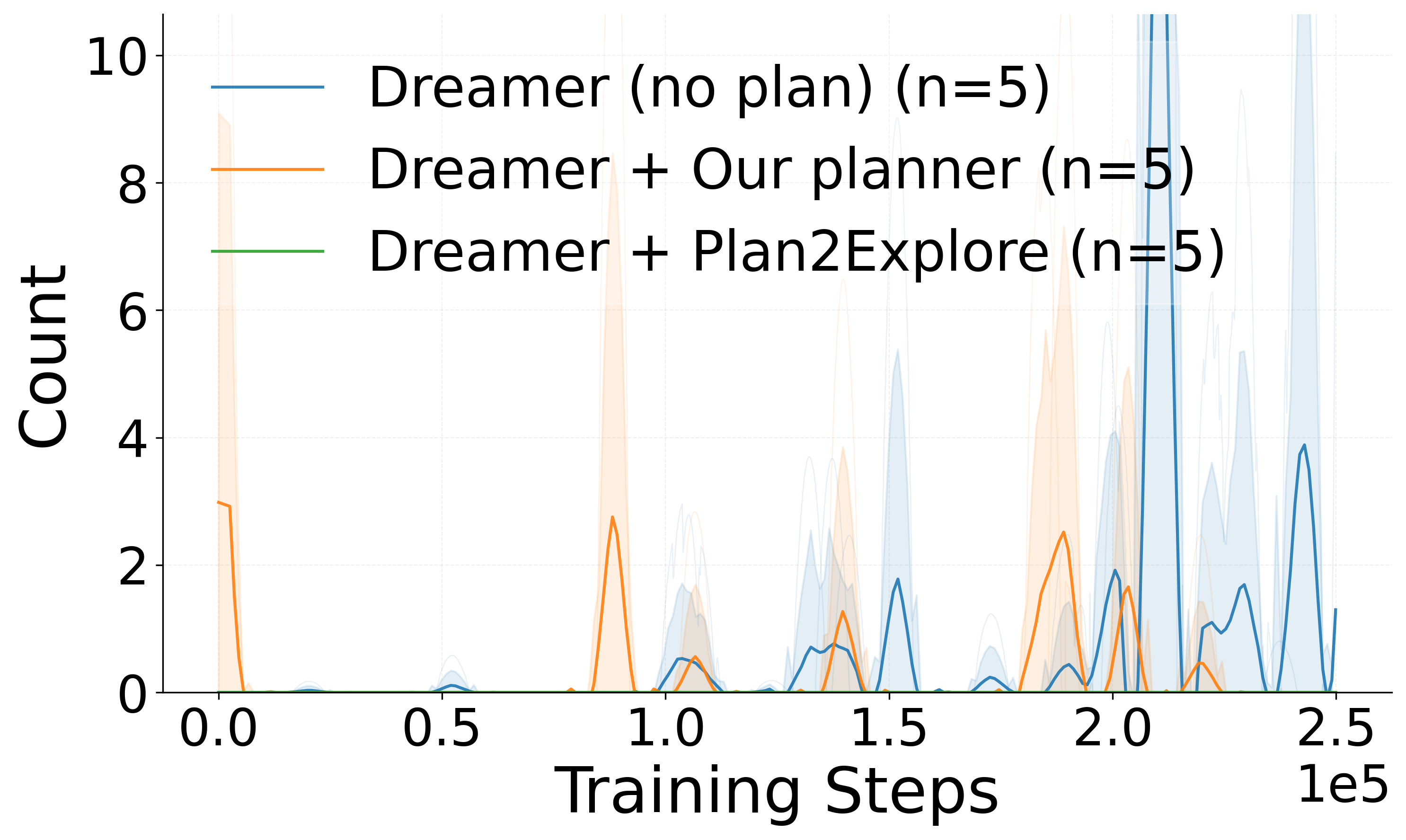}
    \caption{make\_wood\_sword}
    \label{fig:crafter_make_wood_sword}
  \end{subfigure}\hfill
  \begin{subfigure}[t]{0.32\linewidth}
    \centering\includegraphics[width=\linewidth]{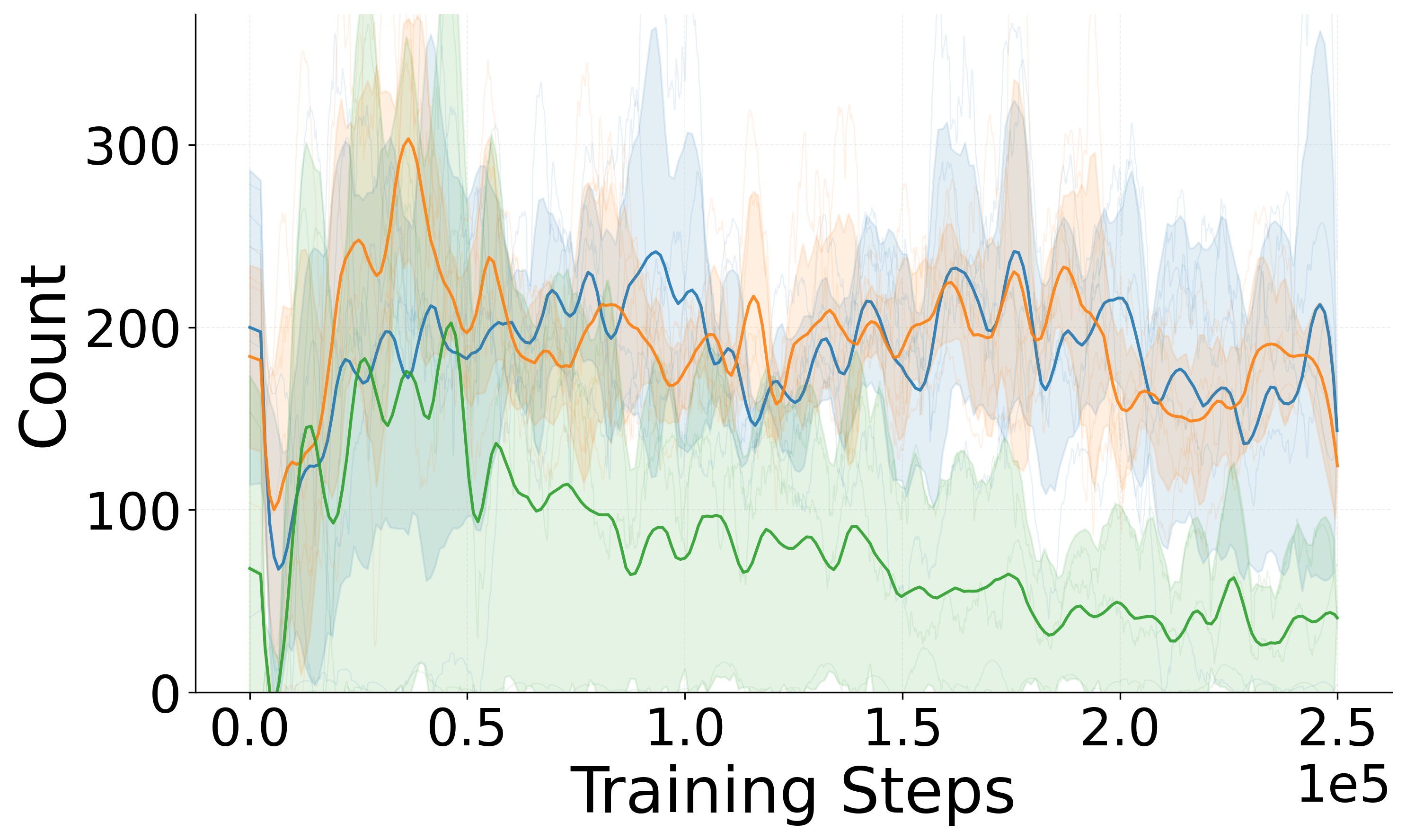}
    \caption{wake\_up}
    \label{fig:crafter_wake_up}
  \end{subfigure}

  \caption{Crafter achievement counts over 300k steps (3 seeds) for no plan vs the planning variant vs plan2explore. The planning agent forms reliable routines (wood, table, zombie) and improves sample efficiency, while deeper crafting remains conservative within this budget.}
  \label{fig:crafter_achievements_grid}
\end{figure*}

To better understand where the return gains in Sec.~\ref{sec:crafter} come from, we report
per-achievement learning curves over the 300k-step budget. Gains are concentrated in routine-forming achievements such as collecting wood, placing tables, and defeating zombies, where short, commit-aware exploratory rollouts appear to reduce dithering and stabilize representation learning (Figs.~\ref{fig:crafter_collect_wood}, \ref{fig:crafter_place_table}, \ref{fig:crafter_defeat_zombie}). In contrast, deeper crafting branches (make wooden sword/pickaxe) remain low within 300k steps, especially for the planning variant (Figs.~\ref{fig:crafter_make_wood_pickaxe}, \ref{fig:crafter_make_wood_sword}). We hypothesise this is because the agent attempts these actions early game and finds they do nothing consistently (a nearby table and correct materials in the inventory are prerequisites before the "make pickaxe" or "make axe" button make the respective tools), so the corresponding states become high-confidence low-reward states and are not attempted again, as detailed in failure mode \#2. Even so, the planning variant remains better than or competitive with the baseline across the panel (Fig.~\ref{fig:crafter_achievements_grid}).

Tasks like collect wood and place table are repeatable and reward-dense; plan commitment converts them into habits, yielding steady slopes and higher returns (Figs.~\ref{fig:crafter_collect_wood}, \ref{fig:crafter_place_table}). Combat against zombies sits between routine and opportunistic: once wood/table routines are established, the planner’s broader coverage increases encounter rate, so the zombie curve rises earlier and higher than no-plan but still exhibits spikes (Fig.~\ref{fig:crafter_defeat_zombie}). Making tools remains low for both agents; the planning variant is especially conservative (Figs.~\ref{fig:crafter_make_wood_pickaxe}, \ref{fig:crafter_make_wood_sword}). It is interesting that even though we do not explicitly optimise for reward in the planner, the inherent bias toward rewarding rollouts results in what is effectively zombie and tree farming behaviour (Figs.~\ref{fig:crafter_collect_wood}, \ref{fig:crafter_defeat_zombie}). Collect drink, collect sapling and eat cow (Figs.~\ref{fig:crafter_collect_drink}, \ref{fig:crafter_collect_sapling}, \ref{fig:crafter_eat_cow}) are all roughly matched between the no plan variant and the planning variant, indicating these routine-light behaviours benefit less from commitment.

\section{Default Configuration and Code Base}
\label{sec:appendix_config}

\subsection{Default Configuration}
The following listing provides the default hyperparameters and settings used in our experiments.

\begin{lstlisting}[basicstyle=\ttfamily\small]
use_plan: True

logdir: null
traindir: null
evaldir: null
offline_traindir: ''
offline_evaldir: ''
seed: 0
deterministic_run: False
steps: 1e6
parallel: False
eval_every: 1e4
eval_episode_num: 10
log_every: 1e4
reset_every: 0
device: 'cuda:0'
compile: True
precision: 16
debug: False

# Environment
task: 'dmc_walker_walk'
size: [64, 64]
# envs: 1
# action_repeat: 1
time_limit: 1000
grayscale: False
prefill: 2500
reward_EMA: True

# Model
dyn_hidden: 512
dyn_deter: 512
dyn_stoch: 32
dyn_discrete: 32
dyn_rec_depth: 2
dyn_mean_act: 'none'
dyn_std_act: 'sigmoid2'
dyn_min_std: 0.1
grad_heads: ['decoder', 'reward', 'cont', 'entropy']
units: 512
act: 'SiLU'
norm: True
encoder:
  {mlp_keys: '$^', cnn_keys: 'image', act: 'SiLU', norm: True, cnn_depth: 64, kernel_size: 4, minres: 4, mlp_layers: 5, mlp_units: 1024, symlog_inputs: True}
decoder:
  {mlp_keys: '$^', cnn_keys: 'image', act: 'SiLU', norm: True, cnn_depth: 32, kernel_size: 4, minres: 4, mlp_layers: 5, mlp_units: 1024, cnn_sigmoid: False, image_dist: mse, vector_dist: symlog_mse, outscale: 1.0}
actor:
  {layers: 2, dist: 'normal', entropy: 3e-4, unimix_ratio: 0.01, std: 'learned', min_std: 0.1, max_std: 1.0, temp: 0.1, lr: 3e-5, eps: 1e-5, grad_clip: 100.0, outscale: 1.0}
Q:
  {layers: 2, dist: 'symlog_disc', slow_target: True, slow_target_update: 1, slow_target_fraction: 0.02, lr: 3e-5, eps: 1e-5, grad_clip: 100.0, outscale: 1.0} 
critic:
  {layers: 2, dist: 'symlog_disc', slow_target: True, slow_target_update: 1, slow_target_fraction: 0.02, lr: 3e-5, eps: 1e-5, grad_clip: 100.0, outscale: 0.0}
reward_head:
  {layers: 2, dist: 'symlog_disc', loss_scale: 1.0, outscale: 1.0} 
entropy_head:
  {layers: 2, dist: 'symlog_disc', loss_scale: 1.0, outscale: 1.0} 
cont_head:
  {layers: 2, loss_scale: 1.0, outscale: 1.0}
dyn_scale: 0.5
rep_scale: 0.1
kl_free: 1.0
weight_decay: 0.0
unimix_ratio: 0.01
initial: 'learned'

# Training
batch_size: 16
batch_length: 64
train_ratio: 512
pretrain: 100
model_lr: 1e-4
opt_eps: 1e-8
grad_clip: 1000
dataset_size: 1000000
opt: 'adam'

# Behavior.
discount: 0.997
discount_lambda: 0.95
imag_horizon: 15
imag_gradient: 'dynamics'
imag_gradient_mix: 0.0
eval_state_mean: False

# Exploration
expl_behavior: 'greedy'
expl_until: 0
expl_extr_scale: 0.0
expl_intr_scale: 1.0
disag_target: 'stoch'
disag_log: True
disag_models: 10
disag_offset: 1
disag_layers: 4
disag_units: 400
disag_action_cond: False

# plan_behavior:
plan_max_horizon: 16
plan_choices: 256
plan_train_every: 32
sub_batch_size: 64
num_epochs: 30
buffer_size: 32768
clip_epsilon: 0.2
gamma: 0.99
lmbda: 0.95
entropy_eps: 0.1
num_cells: 256
lr: 0.003
seq_length: 8
buffer_minimum: 512
meta_action_quant: 5                     # used in CategoricalSpec
num_meta_action_lwr: 2                     # used in CategoricalSpec
ent_multiplier: 1.0                      # multiplier for entropy in _flow method
rew_multiplier: 1.0                    # multiplier for reward in _flow method

dmc_vision:
  steps: 1e6
  action_repeat: 2
  envs: 1
  train_ratio: 512
  video_pred_log: false
  encoder: {mlp_keys: '$^', cnn_keys: 'image'}
  decoder: {mlp_keys: '$^', cnn_keys: 'image'}

crafter:
  task: crafter_reward
  step: 1e6
  action_repeat: 1
  envs: 1
  train_ratio: 512
  video_pred_log: false
  dyn_hidden: 1024
  dyn_deter: 4096
  units: 1024
  encoder: {mlp_keys: '$^', cnn_keys: 'image', cnn_depth: 96, mlp_layers: 5, mlp_units: 1024}
  decoder: {mlp_keys: '$^', cnn_keys: 'image', cnn_depth: 96, mlp_layers: 5, mlp_units: 1024}
  actor: {layers: 5, dist: 'onehot', std: 'none'}
  value: {layers: 5}
  reward_head: {layers: 5}
  cont_head: {layers: 5}
  imag_gradient: 'reinforce'
\end{lstlisting}

\subsection{Code Base}
Our implementation is forked from \url{https://github.com/NM512/dreamerv3-torch/blob/main/dreamer.py}. We adapt this code to our setting while retaining the default configuration listed above.

\end{document}

%% file: math_commands.tex
\usepackage{amsmath,amsfonts,bm}

\def\eqref#1{equation~\ref{#1}}

\def\1{\bm{1}}

\def\ra{{\textnormal{a}}}

\def\ru{{\textnormal{u}}}

\def\rx{{\textnormal{x}}}

\def\rz{{\textnormal{z}}}

\DeclareMathAlphabet{\mathsfit}{\encodingdefault}{\sfdefault}{m}{sl}
\SetMathAlphabet{\mathsfit}{bold}{\encodingdefault}{\sfdefault}{bx}{n}

\newcommand{\KL}{D_{\mathrm{KL}}}



%% file: iclr2026_conference.bib
@article{burda2018exploration,
  title={Exploration by random network distillation},
  author={Burda, Yuri and Edwards, Harrison and Storkey, Amos and Klimov, Oleg},
  journal={arXiv preprint arXiv:1810.12894},
  year={2018}
}

@article{bellemare2016unifying,
  title={Unifying count-based exploration and intrinsic motivation},
  author={Bellemare, Marc and Srinivasan, Sriram and Ostrovski, Georg and Schaul, Tom and Saxton, David and Munos, Remi},
  journal={Advances in neural information processing systems},
  volume={29},
  year={2016}
}

@inproceedings{pathak2017curiosity,
  title={Curiosity-driven exploration by self-supervised prediction},
  author={Pathak, Deepak and Agrawal, Pulkit and Efros, Alexei A and Darrell, Trevor},
  booktitle={International conference on machine learning},
  pages={2778--2787},
  year={2017},
  organization={PMLR}
}

@inproceedings{sekar2020planning,
  title={Planning to explore via self-supervised world models},
  author={Sekar, Ramanan and Rybkin, Oleh and Daniilidis, Kostas and Abbeel, Pieter and Hafner, Danijar and Pathak, Deepak},
  booktitle={International conference on machine learning},
  pages={8583--8592},
  year={2020},
  organization={PMLR}
}

@article{khetarpal2022towards,
  title={Towards continual reinforcement learning: A review and perspectives},
  author={Khetarpal, Khimya and Riemer, Matthew and Rish, Irina and Precup, Doina},
  journal={Journal of Artificial Intelligence Research},
  volume={75},
  pages={1401--1476},
  year={2022}
}

@article{hafner2019dream,
  title={Dream to control: Learning behaviors by latent imagination},
  author={Hafner, Danijar and Lillicrap, Timothy and Ba, Jimmy and Norouzi, Mohammad},
  journal={arXiv preprint arXiv:1912.01603},
  year={2019}
}

@article{hafner2020mastering,
  title={Mastering atari with discrete world models},
  author={Hafner, Danijar and Lillicrap, Timothy and Norouzi, Mohammad and Ba, Jimmy},
  journal={arXiv preprint arXiv:2010.02193},
  year={2020}
}

@article{hafner2023mastering,
  title={Mastering diverse domains through world models},
  author={Hafner, Danijar and Pasukonis, Jurgis and Ba, Jimmy and Lillicrap, Timothy},
  journal={arXiv preprint arXiv:2301.04104},
  year={2023}
}

@article{ecoffet2019go,
  title={Go-explore: a new approach for hard-exploration problems},
  author={Ecoffet, Adrien and Huizinga, Joost and Lehman, Joel and Stanley, Kenneth O and Clune, Jeff},
  journal={arXiv preprint arXiv:1901.10995},
  year={2019}
}

@article{badia2020never,
  title={Never give up: Learning directed exploration strategies},
  author={Badia, Adri{\`a} Puigdom{\`e}nech and Sprechmann, Pablo and Vitvitskyi, Alex and Guo, Daniel and Piot, Bilal and Kapturowski, Steven and Tieleman, Olivier and Arjovsky, Mart{\'\i}n and Pritzel, Alexander and Bolt, Andew and others},
  journal={arXiv preprint arXiv:2002.06038},
  year={2020}
}

@inproceedings{shyam2019model,
  title={Model-based active exploration},
  author={Shyam, Pranav and Ja{\'s}kowski, Wojciech and Gomez, Faustino},
  booktitle={International conference on machine learning},
  pages={5779--5788},
  year={2019},
  organization={PMLR}
}

@book{sutton2018reinforcement,
  author    = {Richard S. Sutton and Andrew G. Barto},
  title     = {Reinforcement Learning: An Introduction},
  edition   = {2nd},
  year      = {2018},
  publisher = {MIT Press},
  url       = {http://incompleteideas.net/book/the-book-2nd.html}
}

@article{quinlan1986induction,
  title={Induction of decision trees},
  author={Quinlan, J. Ross},
  journal={Machine learning},
  volume={1},
  pages={81--106},
  year={1986},
  publisher={Springer}
}

@article{MinigridMiniworld23,
  author       = {Maxime Chevalier-Boisvert and Bolun Dai and Mark Towers and Rodrigo de Lazcano and Lucas Willems and Salem Lahlou and Suman Pal and Pablo Samuel Castro and Jordan Terry},
  title        = {Minigrid \& Miniworld: Modular \& Customizable Reinforcement Learning Environments for Goal-Oriented Tasks},
  journal      = {CoRR},
  volume       = {abs/2306.13831},
  year         = {2023},
}

@article{mppi,
  title={Model predictive path integral control using covariance variable importance sampling},
  author={Williams, Grady and Aldrich, Andrew and Theodorou, Evangelos},
  journal={arXiv preprint arXiv:1509.01149},
  year={2015}
}

@inproceedings{pspic,
  title={Policy search for path integral control},
  author={G{\'o}mez, Vicen{\c{c}} and Kappen, Hilbert J and Peters, Jan and Neumann, Gerhard},
  booktitle={Machine Learning and Knowledge Discovery in Databases: European Conference, ECML PKDD 2014, Nancy, France, September 15-19, 2014. Proceedings, Part I 14},
  pages={482--497},
  year={2014},
  organization={Springer}
}

@article{tdmpc2,
  title={Td-mpc2: Scalable, robust world models for continuous control},
  author={Hansen, Nicklas and Su, Hao and Wang, Xiaolong},
  journal={arXiv preprint arXiv:2310.16828},
  year={2023}
}

@article{tdmpc,
  title={Temporal difference learning for model predictive control},
  author={Hansen, Nicklas and Wang, Xiaolong and Su, Hao},
  journal={arXiv preprint arXiv:2203.04955},
  year={2022}
}

@article{alphago,
  title={Mastering the game of Go with deep neural networks and tree search},
  author={Silver, David and Huang, Aja and Maddison, Chris J and Guez, Arthur and Sifre, Laurent and Van Den Driessche, George and Schrittwieser, Julian and Antonoglou, Ioannis and Panneershelvam, Veda and Lanctot, Marc and others},
  journal={nature},
  volume={529},
  number={7587},
  pages={484--489},
  year={2016},
  publisher={Nature Publishing Group}
}

@article{alphazero,
  title={Mastering chess and shogi by self-play with a general reinforcement learning algorithm},
  author={Silver, David and Hubert, Thomas and Schrittwieser, Julian and Antonoglou, Ioannis and Lai, Matthew and Guez, Arthur and Lanctot, Marc and Sifre, Laurent and Kumaran, Dharshan and Graepel, Thore and others},
  journal={arXiv preprint arXiv:1712.01815},
  year={2017}
}

@inproceedings{ostrovski2017count,
  title     = {Count-Based Exploration with Neural Density Models},
  author    = {Ostrovski, Georg and Bellemare, Marc G. and van den Oord, A{\"a}ron and Munos, R{\'e}mi},
  booktitle = {Proceedings of the 34th International Conference on Machine Learning},
  pages     = {2721--2730},
  year      = {2017},
  organization = {PMLR}
}

@inproceedings{burda2019largescale,
  title     = {Large-Scale Study of Curiosity-Driven Learning},
  author    = {Burda, Yuri and Edwards, Harri and Pathak, Deepak and Storkey, Amos and Darrell, Trevor and Efros, Alexei A.},
  booktitle = {International Conference on Learning Representations},
  year      = {2019}
}

@inproceedings{jarrett2023hindsight,
  title     = {Curiosity in Hindsight: Intrinsic Exploration in Stochastic Environments},
  author    = {Jarrett, Daniel and Tallec, Corentin and Altch{\'e}, Florent and Mesnard, Thomas and Munos, Remi and Valko, Michal},
  booktitle = {Proceedings of the 40th International Conference on Machine Learning},
  pages     = {14780--14816},
  year      = {2023},
  organization = {PMLR}
}

@article{mahankali2022novelty,
  title   = {Does Novelty-Based Exploration Maximize Novelty?},
  author  = {Mahankali, Srinath and Hong, Zhang-Wei and Agrawal, Pulkit},
  journal = {arXiv preprint arXiv:2211.07627},
  year    = {2022}
}

@inproceedings{mcts,
  title={Efficient selectivity and backup operators in Monte-Carlo tree search},
  author={Coulom, R{\'e}mi},
  booktitle={International conference on computers and games},
  pages={72--83},
  year={2006},
  organization={Springer}
}

@article{Shannon1948,
  author  = {Shannon, Claude E.},
  title   = {A Mathematical Theory of Communication},
  journal = {Bell System Technical Journal},
  volume  = {27},
  number  = {3},
  pages   = {379--423},
  year    = {1948}
}

@misc{MarshContinuousEntropy,
  author       = {Marsh, Charles R.},
  title        = {An Introduction to Continuous Entropy},
  howpublished = {\url{https://charlesmarsh.com/continuous-entropy/}},
  note         = {Accessed: 2025-12-02},
  year         = {2013}
}

@article{chua2018deep,
  title={Deep reinforcement learning in a handful of trials using probabilistic dynamics models},
  author={Chua, Kurtland and Calandra, Roberto and McAllister, Rowan and Levine, Sergey},
  journal={Advances in neural information processing systems},
  volume={31},
  year={2018}
}

@article{raileanu2020ride,
  title={Ride: Rewarding impact-driven exploration for procedurally-generated environments},
  author={Raileanu, Roberta and Rockt{\"a}schel, Tim},
  journal={arXiv preprint arXiv:2002.12292},
  year={2020}
}

@inproceedings{oc,
  title={The option-critic architecture},
  author={Bacon, Pierre-Luc and Harb, Jean and Precup, Doina},
  booktitle={Proceedings of the AAAI conference on artificial intelligence},
  volume={31},
  number={1},
  year={2017}
}

@article{hippo,
  title={Sub-policy adaptation for hierarchical reinforcement learning},
  author={Li, Alexander C and Florensa, Carlos and Clavera, Ignasi and Abbeel, Pieter},
  journal={arXiv preprint arXiv:1906.05862},
  year={2019}
}

@article{med,
  title={Maximum entropy model-based reinforcement learning},
  author={Svidchenko, Oleg and Shpilman, Aleksei},
  journal={arXiv preprint arXiv:2112.01195},
  year={2021}
}

@article{raif,
  title={R-aif: Solving sparse-reward robotic tasks from pixels with active inference and world models},
  author={Nguyen, Viet Dung and Yang, Zhizhuo and Buckley, Christopher L and Ororbia, Alexander},
  journal={arXiv preprint arXiv:2409.14216},
  year={2024}
}

@inproceedings{lbyl,
  title={Look before you leap: Bridging model-free and model-based reinforcement learning for planned-ahead vision-and-language navigation},
  author={Wang, Xin and Xiong, Wenhan and Wang, Hongmin and Wang, William Yang},
  booktitle={Proceedings of the European Conference on Computer Vision (ECCV)},
  pages={37--53},
  year={2018}
}

@article{mgo,
  title={Mastering the game of go without human knowledge},
  author={Silver, David and Schrittwieser, Julian and Simonyan, Karen and Antonoglou, Ioannis and Huang, Aja and Guez, Arthur and Hubert, Thomas and Baker, Lucas and Lai, Matthew and Bolton, Adrian and others},
  journal={nature},
  volume={550},
  number={7676},
  pages={354--359},
  year={2017},
  publisher={Nature Publishing Group UK London}
}

@article{drone,
  title={Champion-level drone racing using deep reinforcement learning},
  author={Kaufmann, Elia and Bauersfeld, Leonard and Loquercio, Antonio and M{\"u}ller, Matthias and Koltun, Vladlen and Scaramuzza, Davide},
  journal={Nature},
  volume={620},
  number={7976},
  pages={982--987},
  year={2023},
  publisher={Nature Publishing Group UK London}
}

@article{switch1,
  title={Learnings options end-to-end for continuous action tasks},
  author={Klissarov, Martin and Bacon, Pierre-Luc and Harb, Jean and Precup, Doina},
  journal={arXiv preprint arXiv:1712.00004},
  year={2017}
}

@article{switch2,
  title={Attention option-critic},
  author={Chunduru, Raviteja and Precup, Doina},
  journal={arXiv preprint arXiv:2201.02628},
  year={2022}
}

@article{switch3,
  title={Hierarchical Reinforcement Learning in Multi-Goal Spatial Navigation with Autonomous Mobile Robots},
  author={Johnson, Brendon and Weitzenfeld, Alfredo},
  journal={arXiv preprint arXiv:2504.18794},
  year={2025}
}

@article{world_models,
  title={World models},
  author={Ha, David and Schmidhuber, J{\"u}rgen},
  journal={arXiv preprint arXiv:1803.10122},
  year={2018}
}

@article{schulman2017ppo,
  title        = {Proximal Policy Optimization Algorithms},
  author       = {Schulman, John and Wolski, Filip and Dhariwal, Prafulla and Radford, Alec and Klimov, Oleg},
  journal      = {arXiv preprint arXiv:1707.06347},
  year         = {2017}
}

@inproceedings{hubert2021learning,
  title={Learning and planning in complex action spaces},
  author={Hubert, Thomas and Schrittwieser, Julian and Antonoglou, Ioannis and Barekatain, Mohammadamin and Schmitt, Simon and Silver, David},
  booktitle={International Conference on Machine Learning},
  pages={4476--4486},
  year={2021},
  organization={PMLR}
}

@article{kaelbling1998pomdp,
  title   = {Planning and acting in partially observable stochastic domains},
  author  = {Kaelbling, Leslie Pack and Littman, Michael L. and Cassandra, Anthony R.},
  journal = {Artificial Intelligence},
  volume  = {101},
  number  = {1--2},
  pages   = {99--134},
  year    = {1998},
  doi     = {10.1016/S0004-3702(98)00023-X}
}

@inproceedings{antonoglou2021planning,
  title={Planning in stochastic environments with a learned model},
  author={Antonoglou, Ioannis and Schrittwieser, Julian and Ozair, Sherjil and Hubert, Thomas K and Silver, David},
  booktitle={International Conference on Learning Representations},
  year={2021}
}

@book{CoverThomas2006,
  title={Elements of Information Theory},
  author={Cover, Thomas M. and Thomas, Joy A.},
  year={2006},
  publisher={Wiley}
}

@inproceedings{engstrom2020implementation,
  title        = {Implementation Matters in Deep Policy Gradients: A Case Study on {PPO} and {TRPO}},
  author       = {Engstrom, Logan and Ilyas, Andrew and Santurkar, Shibani and Tsipras, Dimitris and Janoos, Firdaus and Rudolph, Larry and Madry, Aleksander},
  booktitle    = {International Conference on Learning Representations (ICLR)},
  year         = {2020},
  note         = {arXiv:2005.12729}
}

@inproceedings{henderson2018matters,
  title        = {Deep Reinforcement Learning that Matters},
  author       = {Henderson, Peter and Islam, Riashat and Bachman, Philip and Pineau, Joelle and Precup, Doina and Meger, David},
  booktitle    = {AAAI Conference on Artificial Intelligence (AAAI)},
  year         = {2018},
  doi          = {10.1609/aaai.v32i1.11694}
}

@article{kostrikov2020image,
  title        = {Image Augmentation Is All You Need: Regularizing Deep Reinforcement Learning from Pixels},
  author       = {Kostrikov, Ilya and Yarats, Denis and Fergus, Rob},
  journal      = {arXiv preprint arXiv:2004.13649},
  year         = {2020}
}

@article{yarats2021drqv2,
  title        = {Mastering Visual Continuous Control: Improved Data-Augmented Reinforcement Learning},
  author       = {Yarats, Denis and Fergus, Rob and Lazaric, Alessandro and Pinto, Lerrel},
  journal      = {arXiv preprint arXiv:2107.09645},
  year         = {2021}
}

@article{laskin2020rad,
  title        = {Reinforcement Learning with Augmented Data},
  author       = {Laskin, Michael and Lee, Kimin and Stooke, Adam and Pinto, Lerrel and Abbeel, Pieter and Srinivas, Aravind},
  journal      = {arXiv preprint arXiv:2004.14990},
  year         = {2020},
  note         = {NeurIPS 2020 camera-ready}
}

@inproceedings{srinivas2020curl,
  title        = {{CURL}: Contrastive Unsupervised Representations for Reinforcement Learning},
  author       = {Srinivas, Aravind and Laskin, Michael and Abbeel, Pieter},
  booktitle    = {Proceedings of the 37th International Conference on Machine Learning (ICML)},
  year         = {2020},
  publisher    = {PMLR},
  url          = {https://proceedings.mlr.press/v119/laskin20a.html}
}

@article{crafter,
  title   = {Benchmarking the Spectrum of Agent Capabilities},
  author  = {Danijar Hafner},
  journal = {arXiv preprint arXiv:2109.06780},
  year    = {2021},
  url     = {https://arxiv.org/abs/2109.06780}
}

@article{dmc,
  title   = {DeepMind Control Suite},
  author  = {Yuval Tassa and Yotam Doron and Alistair Muldal and Tom Erez and Yazhe Li and Diego de Las Casas and David Budden and Abbas Abdolmaleki and Josh Merel and Andrew Lefrancq and Timothy P. Lillicrap and Martin A. Riedmiller},
  journal = {arXiv preprint arXiv:1801.00690},
  year    = {2018},
  url     = {https://arxiv.org/abs/1801.00690}
}
